\documentclass[10pt,twocolumn,letterpaper]{article}
\usepackage[accsupp]{axessibility} 
\usepackage{authblk}
\usepackage{cvpr} 
\usepackage{float}
\usepackage{graphicx}
\usepackage{amsmath}
\usepackage{amssymb}
\usepackage{indentfirst} 
\usepackage{booktabs}
\usepackage{subcaption}
\usepackage{multirow}
\usepackage{diagbox}
\usepackage{stfloats}
\usepackage{makecell}
\usepackage{graphicx}
\usepackage{appendix}
\usepackage{colortbl}
\usepackage[dvipsnames]{xcolor}

\definecolor{cvprblue}{rgb}{0.21,0.49,0.74}
\usepackage[pagebackref,breaklinks,colorlinks,citecolor=cvprblue]{hyperref}
\def\paperID{13778} 
\def\confName{CVPR}
\def\confYear{2024}
\newcommand*\samethanks[1][\value{footnote}]{\footnotemark[#1]}

\title{Learning Transferable Negative Prompts for Out-of-Distribution Detection}

\author[1]{Tianqi Li}
\author[2]{Guansong Pang\thanks{Corresponding author: G. Pang (gspang@smu.edu.sg) and X. Bai (baixiao@buaa.edu.cn)}}
\author[1]{Xiao Bai\samethanks}
\author[1]{Wenjun Miao}
\author[1]{Jin Zheng}
\affil[1]{School of Computer Science and Engineering, State Key Laboratory of Complex \& Critical Software Environment, Jiangxi Research Institute, Beihang University, China}
\affil[2]{School of Computing and Information Systems, Singapore Management University}

\begin{document}
\maketitle
\begin{abstract}

Existing prompt learning methods have shown certain capabilities in Out-of-Distribution (OOD) detection, but the lack of OOD images in the target dataset in their training can lead to mismatches between OOD images and In-Distribution (ID) categories, resulting in a high false positive rate. To address this issue, we introduce a novel OOD detection method, named `NegPrompt', to learn a set of negative prompts, each representing a negative connotation of a given class label, for delineating the boundaries between ID and OOD images. It learns such negative prompts with ID data only, without any reliance on external outlier data.
Further, current methods assume the availability of samples of all ID classes, rendering them ineffective in open-vocabulary learning scenarios where the inference stage can contain novel ID classes not present during training. In contrast, our learned negative prompts are transferable to novel class labels.
Experiments on various ImageNet benchmarks show that NegPrompt surpasses state-of-the-art prompt-learning-based OOD detection methods and maintains a consistent lead in hard OOD detection in closed- and open-vocabulary classification scenarios. Code is available at \renewcommand\UrlFont{\color{blue}}\url{https://github.com/mala-lab/negprompt}.

\vspace{-0.5cm}
\end{abstract}    
\section{Introduction}
\label{sec:intro}

Since the advent of deep learning, numerous image recognition models~\cite{he2016deep, liu2021swin, dosovitskiy2020image} have relied solely on image features for classification. However, in recent years large pre-trained vision-language models (VLMs), such as CLIP~\cite{radford2021learning}, integrated natural language processing into computer vision, enhancing the semantic understanding capabilities of computer vision models. It has been observed that these VLMs excel in image classification, particularly in zero-shot scenarios. This is attributed to their extensive self-supervised pre-training on web-scale image-text data, which has endowed them with robust semantic transfer abilities.

Despite the strong zero-shot classification capabilities of VLMs, numerous research efforts are put to unleash their potential, \eg, by investigating whether models like CLIP can achieve enhanced performance with training on downstream target datasets. This has led to the development of various techniques to fine-tune CLIP~\cite{gao2023clip, jia2022visual}, among which prompt learning has sparked widespread interest. Prompt learning (or prompt tuning)~\cite{shin2020autoprompt,jiang2020can,zhong2021factual}, a methodology originating from natural language processing, focuses on learning the prompt inputs into a large-scale pre-trained network, rather than learning or fine-tuning the parameters of the network. In CLIP, a common prompt is `\texttt{a photo of a [class name]}'. The aim of prompt learning, \eg, in approaches like CoOp \cite{zhou2022learning}, is to learn a soft/differentiable context vector to replace the fixed text prompt like `\texttt{a photo of a}', thereby leveraging CLIP's powerful generalization in semantic understanding while also fine-tuning for specific target datasets~\cite{zhou2022learning,zhou2022conditional}.

\begin{figure}[t]

  \centering
    \includegraphics[width=1\linewidth]{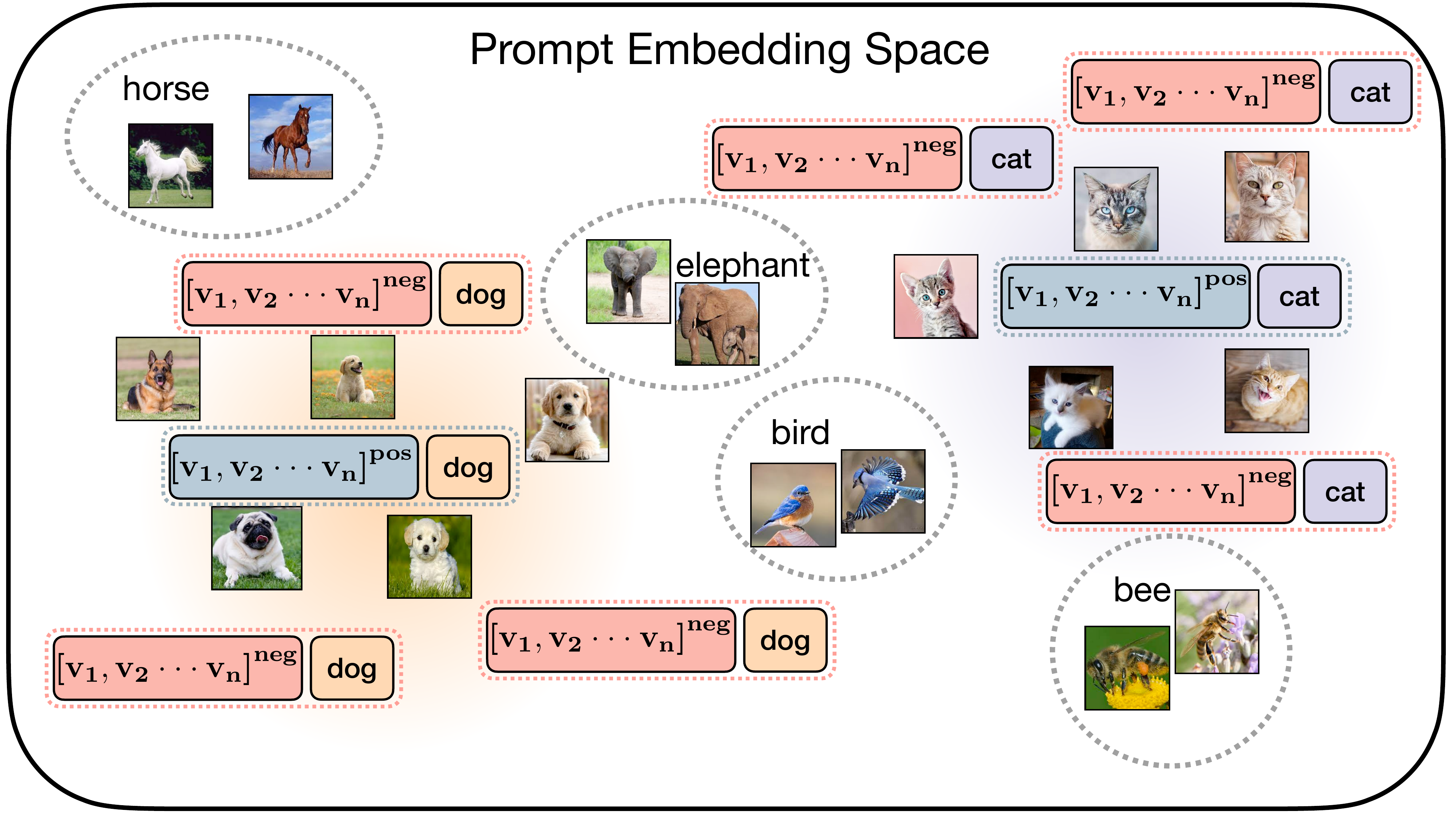}
  \caption{Illustration of the key intuition of NegPrompt. For each ID class, NegPrompt trains a small set of learnable prompts that have negative semantics to the learned positive prompt of the given class. As a result, OOD samples exhibit higher similarity to the negative prompts than the positive prompts.}

  \label{fig:motivation}
  \vspace{-0.5cm}
\end{figure}

However, even though prompt learning enhances the target dataset perception capabilities of VLMs, it struggles with Out-Of-Distribution (OOD) detection~\cite{hendrycks2017a, liang2018enhancing}. Under the OOD detection task, the test set comprises images from both the training classes -- in-distribution (ID) data -- and images from other unknown categories (OOD images). VLM-based classification is typically performed by first replacing the `\texttt{[class name]}' with the class label of each category in a prompt, which is then processed by the text encoder of CLIP to obtain the class embedding and assign the class label that has the highest cosine similarity with the embedding of a test image from the image encoder of CLIP. However in OOD detection tasks, the models do not have access to the class names of OOD images, thereby lacking the knowledge about OOD data.

This issue, exacerbated by the models' tendency towards overconfidence that often predicts OOD data as ID images with high confidence~\cite{nguyen2015deep}, undermines their ability to effectively detect OOD images.

There have been a number of CLIP-based methods designed specifically for VLM-driven OOD detection methods, but most of them~\cite{esmaeilpour2022zero, wang2023clipn, ming2022delving} focused on a zero-shot setting where no training data of the target dataset is available. Due to the lack of adaptation to the target dataset, they tend to detect unusual ID images as OOD data. Among them, CLIPN \cite{wang2023clipn}, which trains an additional `no' text encoder to provide prompt embeddings of not having specific classes, is the best detector, but it relies on a large-scale auxiliary dataset to train such a text encoder and it is computationally expensive.

The most related work is a very recent approach LoCoOp~\cite{miyai2023locoop} that utilizes the training ID data to tune CLIP to capture local features of the ID classes in the prompt. It shows substantially improved performance compared to zero-shot methods \cite{ming2022delving}, but it may compromise the ID classification accuracy due to an overemphasis on modeling local features. LoCoOp also lacks knowledge about OOD samples, making it difficult to differentiate boundary ID/OOD images.

In this work, we propose a novel CLIP-based OOD detection method, named \textbf{NegPrompt}. Inspired by \cite{LI2024110258}, Negprompt is designed to learn a set of \textit{negative prompts}, each representing a negative connotation of a given ID class label, to delineate the boundaries between ID and OOD images, as shown in Fig. \ref{fig:motivation}. The negative prompts represent specialized ID-class-dependent concepts, guiding the models to pay attention to the characteristics that are contrary to or disjoint from the ID classes. NegPrompt aims to utilize ID training data and \textit{positive prompts} (\ie, the text prompt embeddings of ID classes) to learn such negative prompts in a way to which OOD images exhibits higher similarity than ID images. 

Essentially, the learned negative prompts have similar semantics to the text prompts generated from the `no' text encoder in CLIPN, but NegPrompt presents a fundamentally different approach: it capitalizes on the generalization ability of the CLIP model and learns the negative prompts with training ID data only, eliminating the reliance on external data and the extensive computation overhead as in CLIPN.

Furthermore, benefiting from the superior generalization ability of CLIP, our method does not require the exposure to all ID classes during training. In other words, the model learns \textit{transferable negative prompts} by using only a small subset of the ID classes, after which we can obtain the negative prompts for the other ID classes by simply replacing the `\texttt{[class name]}' in the prompts with the name of those unexposed classes. This allows our model to work in open-vocabulary~\cite{minderer2022simple, fang2023simple} learning settings, where the models are required to classify images of novel classes that are not seen during training, in addition to a set of training base classes.

Our main contributions can be summarized as follows:

 \begin{itemize}
     \item We propose a prompt learning-based OOD detection approach NegPrompt, which is able to learn negative semantics relative to specific ID classes, thereby enhancing the VLMs' sensitivity to unknown samples. It is a lightweight method that does not require training extra encoders on external data as in related methods \cite{wang2023clipn}.
     \item NegPrompt possesses an open-vocabulary capability due to the transferability of its negative prompts. This means that with training images from just a small subset of ID classes and the class names of all IDs, we can achieve OOD detection on test data with all these ID classes. To the best of our knowledge, there have been no previous fine-tuning methods exploring such a capability.
     \item Extensive experiments on multiple ImageNet-based benchmarks  show that NegPrompt consistently outperforms current state-of-the-art methods in both conventional and hard OOD Detection settings.

 \end{itemize}

\vspace{-0.2cm}
\section{Related Work}

\subsection{Pre-trained Vision-Language Models}
\vspace{-0.2cm}
Understanding the semantic information of images remains a significant challenge in the field of computer vision. With the advent of Transformer~\cite{vaswani2017attention} in computer vision tasks~\cite{dosovitskiy2020image}, CLIP~\cite{radford2021learning} has been introduced as one of the most advanced pre-trained VLMs. Utilizing contrastive learning~\cite{khosla2020supervised}, large-scale models and datasets~\cite{schuhmann2022laion}, CLIP employs image-text pairs as the training data for self-supervised learning. This approach has successfully trained the model to align visual and text signals in a latent space. Concurrently, other researchers~\cite{alayrac2022flamingo, li2022blip, zhang2023adding} have also shown the remarkable generalization capabilities of CLIP and similar vision-language models~\cite{jia2021scaling} in various downstream tasks~\cite{fang2023simple,zhu2024toward,khattak2023maple}.

\subsection{Prompt Learning}
\vspace{-0.2cm}

The concept of prompt learning was initially focused on automating the creation of templates/prompts for extracting knowledge from Bert~\cite{devlin2018bert} or GPT~\cite{radford2018improving}. To bypass manual creation, prompt learning advocates the use of supervised learning to automate the prompt development.~\cite{shin2020autoprompt} proposed a gradient-based approach for identifying the best prompt, establishing the foundation of prompt learning. Later, CoOp~\cite{zhou2022learning} integrated prompt learning into computer vision. CoOp learns a segment of the context before it is fed into the text encoder of CLIP, thus tailoring the learned prompt to the specific target dataset. Many other prompt learning methods for different vision tasks~\cite{zhou2022conditional,jia2022visual,zhou2023anomalyclip,wu2023vadclip,khattak2023maple,sun2022dualcoop,hu2023dualcoop++} are subsequently introduced. However, they are not designed for OOD detection, so they struggle with dealing with the unknown OOD samples during inference.

\subsection{Out-of-Distribution Detection}
\vspace{-0.2cm}
OOD detection is committed to identifying images in image classification tasks that belong to categories not present in the training dataset, typically originating from a different distribution. While traditional OOD detection methods often tackle the problem by either exploiting the prediction logits to define OOD scores~\cite{hendrycks2017a, liang2018enhancing, liu2020energy,hendrycks2019scaling, huang2021importance} or focusing on the class-agnostic information in feature space that is not recoverable from logits~\cite{sun2021react, 9879414}, recent methods~\cite{hendrycks2018deep, kong2021opengan, zhou2021learning, huang2021mos,tian2022pixel,liu2023residual,miao2023out} introduce extra or synthetic OOD data, employing fine-tuning to elevate their model's sensitivity towards unknown classes.

With the introduction of large pre-trained VLMs, OOD detection has embarked on a new trajectory driven by VLMs. MCM~\cite{ming2022delving} aims to integrate the idea of maximum softmax probability \cite{hendrycks2017a} into the inference process of CLIP, while ZOC \cite{esmaeilpour2022zero} enhances OOD detection in a zero-shot setting by learning an additional image interpreter and guessing the category of images. CLIPN~\cite{wang2023clipn} and LoCoOp~\cite{miyai2023locoop}, the most related methods to ours, are based on text prompts. However, CLIPN, during its pre-training phase, trains an additional negative text encoder using a large external dataset to improve its negative semantic prompt, which increases network parameters and deviates from prompt learning that is focused on tuning the target data. LoCoOp, on the other hand, uses prompt learning for matching text and image local features, which can compromise the global perception capability of CLIP and reduce classification accuracy for in-distribution (ID) samples. Also, its model lacks knowledge about OOD samples, which can often lead to high detection errors.

\begin{figure*}[t]
  \centering
  \setlength{\abovecaptionskip} {0.cm}
  \includegraphics[width=0.82\linewidth]{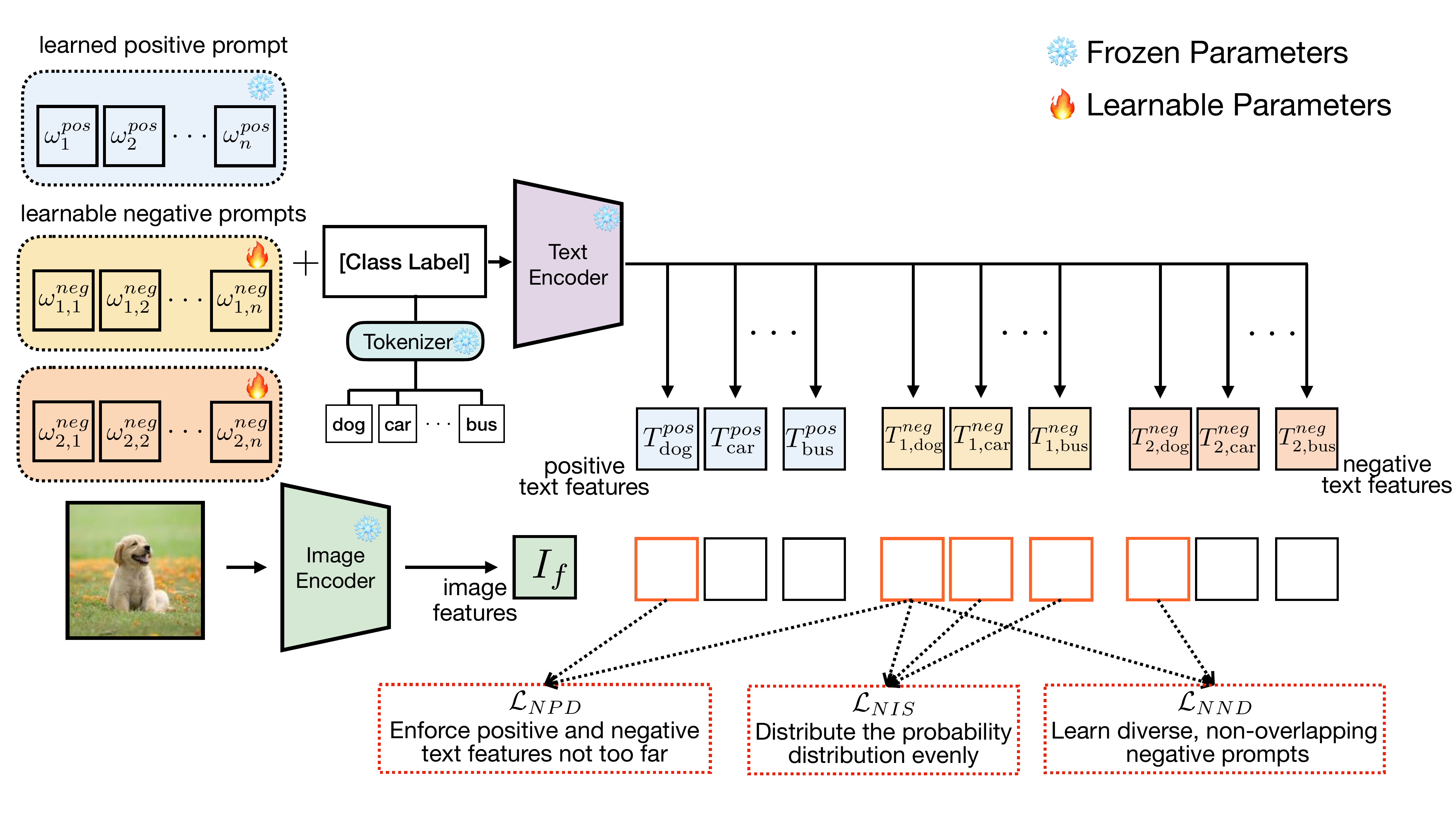}
  \vspace{-0.2cm}
  \caption{Overview of NegPropmt. Given the CLIP model and positive prompts learned by existing prompt learning methods such as CoOp \cite{zhou2022learning}, NegPrompt learns a set of negative prompts relative to different ID class labels via three loss functions that enforce the separation between negative prompts and ID images, and between negative and positive prompts, as well as the diversity of the negative prompts.}

  \label{fig:method_fig}
   \vspace{-0.3cm}
\end{figure*}

\section{Method}
\vspace{-0.2cm}
In this paper, we propose an approach named NegPropmt that leverages pre-trained VLMs, specifically CLIP, to learn negative prompts relative to ID classes for the purpose of OOD detection. The negative prompts are learned in the CLIP's text-image-aligned embedding space with the support of training ID data and their positive prompts (\ie, prompt embeddings of ID classes); no external outlier data is required. Due to its general effectiveness, the popular prompt learning method CoOp \cite{zhou2022learning} is used by default to provide the positive prompts for training NegPrompt.

\subsection{Preliminaries}
\vspace{-0.1cm}

\noindent\textbf{Problem Statement.} Formally, we assume that we have two datasets, namely ID dataset denoted as $D^{in}$ and OOD dataset denoted as $D^{out}$. The ID dataset consists of image-label pairs $(x^{in}, y^{in})$, where $y^{in} \in Y^{in} = \{0, 1, 2, 3...k\}$ belong to the ID class set. Similarly, the OOD dataset contains image-label pairs $(x^{out}, y^{out})$, but all $y^{out} \in Y^{out}=\{ k+1, k+2, ...\}$ belong to the OOD class set. It is important to note that these two sets do not intersect, meaning that $Y^{in} \cap Y^{out} = \varnothing$. As we have a test set $X^{test}$ consists of images from ID and OOD, the goal of OOD detection is to train a classifier $\phi(x)$ that takes an image $x$ as input and returns whether the image belongs to OOD. Unlike existing studies using full-/zero-shot ID training samples, we only use a few samples (16 samples per class) for $D^{in}_{train}$ like \cite{miyai2023locoop} does. In the open-vocabulary detection setting, we only use a small part (10\%) but not all the classes of $D^{in}_{train}$.

\noindent\textbf{CLIP and CoOp.} CLIP~\cite{radford2021learning} is currently one of the most popular image-text models. During the pre-training phase, it uses large-scale image-text pairs for self-supervised contrastive learning, aligning images and texts into the same latent space. The main components of CLIP are an image encoder $Encoder^{image}(I)$ and a text encoder $Encoder^{text}(T)$, which respectively accept image and text inputs. In zero-shot image classification tasks, assume we have $k$ class labels for classification, such as ``cat'', ``dog'', etc., CLIP first incorporates the class labels into pre-designed hard/unlearnable text prompts, such as ``\texttt{a photo of a [class name]}", forming a prompt input set of ``\texttt{a photo of a cat}'', ``\texttt{a photo of a dog}'' and so on. These prompts are then individually fed into the text encoder $Encoder^{text}(T)$ to obtain $k$ text features $T^f$. The testing image is then input into the image encoder to obtain an image feature $I^f$. The cosine similarity is calculated between the normalized image feature and all text features, formally, $Sim(T^f, I^f) = T^f \cdot I^f$, and the text feature $T^f$ with the highest similarity to $I^f$ is considered to be the category to which the image belongs.

Besides zero-shot classification, many have explored ways to improve CLIP’s performance when the target data is accessible. CoOp~\cite{zhou2022learning} introduces prompt learning with CLIP by freezing its encoders and using backpropagation to learn dataset-specific soft/learnable prompts. The text prompts are represented as $t_i = \{\omega^{pos}_1, \omega^{pos}_2, ..., \omega^{pos}_n, c_i\}$, where $c_i$ is the word embedding of the class name and $\omega^{pos}$s are learnable vectors that have positive semantics w.r.t. the class $i$. The goal is to optimize these $\omega^{positive}$s. Specifically, $t_i$ is processed by $Encoder^{text}$ to yield $T^{f,pos}_i$ as a positive prompt embedding of the class $i$, and then the prediction probability is computed as:
\begin{equation}\label{eqn:softmax}
p(y=i | x)=\frac{\exp (sim(T^{f, pos}_i, I^f) / \tau)}{\sum_{j=1}^k \exp (sim(T^{f,pos}_j, I^f) / \tau)},
\end{equation}
where $I^f = Encoder^{image}(x)$ and $\tau$ is a temperature parameter. Next, cross-entropy loss is used to maximize similarity between ID class text embeddings and images.
\begin{equation}
\mathcal{L}_{positive} = \mathbb{E}_{\mathbf{x}_{\text {in }} \sim D_{\text {train }}^{\text {in }}}[-\log(p(y=i | x)].
\end{equation}
This method largely improves CLIP's classification performance by adapting the learnable prompts to the target data. The resulting prompt embeddings of ID classes are used as positive prompts to support the accurate learning of negative prompts in our method below.

\subsection{Proposed Approach}
\vspace{-0.2cm}

NegPrompt aims to learn a set of negative prompts, each representing a negative connotation of an ID class. As shown in Fig. \ref{fig:method_fig}, it involves generating a series of prompts that resemble the positive prompts obtained through CoOp, but with a negative class semantic, such as ``A photo of not a [class label]''. These negative prompts, combined with class labels, can generate a range of negative text features around the positive prompts, so that OOD images exhibit higher similarity to the negative prompts than ID images.

\subsubsection{Learning Negative Prompts}
\vspace{-0.2cm}
To acquire accurate negative prompts representing negative class semantics, we first utilize CoOp to learn positive prompts $\omega^{pos}$s, after which we consider positive prompts to accurately capture the class semantics of samples in the ID dataset. Thereafter, we freeze the positive prompts and focus solely on learning the negative prompts. 
A negative prompt is denoted as $\{\omega^{neg}_1, \omega^{neg}_2, ..., \omega^{neg}_n\}$, where $n$ represents the number of context vectors we aim to learn. By utilizing the frozen CLIP text encoder, we can obtain negative prompt embeddings $T^{f,neg}_{l,i} = Encoder^{text}(t^{neg}_{l,i})$, where $t^{neg}_{l,i}=\{\omega^{neg}_{l,1}, \omega^{neg}_{l,2}, ..., \omega^{neg}_{l,n},c_i\}$ is the $l$-th negative prompt relative the ID class $i$. Thus, due to the presence of the negative prompts, we turn the prediction probability into the following form:
\begin{equation}
\label{eq:softmax}
\begin{split}
    p(y=i | x)= \frac{\exp (S^{f, pos}_i)}{\sum_{j=1}^k \exp (S^{f, pos}_j) + \sum_{l=1}^p\sum_{j=1}^k \exp (S^{f, neg}_{l,j})}
\end{split}
\end{equation}
where $S^{f, pos}_j=sim(T^{f,pos}_{j}, I^f)/\tau$, $S^{f, neg}_{l,j}=sim(T^{f,neg}_{l,j}, I^f)/\tau$ and $p$ is the number of negative prompts we aim to learn for each ID class.
The objective is to learn negative text prompt embeddings that can effectively separate ID and OOD samples around the positive text features. To achieve this, we introduce the following three loss functions:

\noindent\textbf{Negative-Image Separation Loss.}  One of our objectives is to have the negative text prompt embeddings serve as the closest text features to OOD images. However, we only have images from the ID dataset; no OOD images are available. To remedy this problem, we take an alternative approach that aims to push the negative text features away from the ID images. In this regard, we draw inspiration from OE~\cite{hendrycks2018deep}. Contrary to the approach in OE, where the probability distribution of outlier images is evenly distributed among all ID labels, we distribute the probability distribution obtained from ID images evenly among all negative prompts. Since the network parameters are frozen, this drives the negative text features to move away from the ID images, resulting in the learning of prompts with a negative connotation. The formulation is as follows:
\begin{equation}
    \mathcal{L}_{NIS} = \mathbb{E}_{x_{\text {in }} \sim D_{\text {train }}^{\text {in }}}[H(\mathbf{u} ; F(x))],
\end{equation}
where $F(x)$ is the probability vector computed as $Softmax(S^{f,neg})$, and $\mathbf{u}$ is a uniform distribution and $H$ is the cross entropy loss.

\noindent\textbf{Negative-Positive Distance Loss.} To avoid learning trivial negative prompts that are distant from both ID and OOD images, we need to control the negative text feature within a certain range between the ID and OOD images. To this end, we devise a constraint, enforcing that the negative text feature does not deviate too far from the positive text feature. Therefore, we introduce the following loss function to guarantee a certain level of similarity between the negative and positive text feature in the latent space:
\begin{equation}
    \mathcal{L}_{NPD} = -\frac{1}{k*p}\sum_{j=1}^{k}\sum_{i=1}^{p}sim(T^{f,neg}_{i,j}, T^{f, pos}_{j}).
\end{equation}

\noindent\textbf{Negative-Negative Distance Loss.} Furthermore, to ensure that we are effectively learning diverse, non-overlapping negative prompts, we extend the distances between different negative text features within the same label via the following loss function:
\begin{equation}
    \mathcal{L}_{NND} = \frac{1}{k*p*(p-1)}\sum_{j=1}^{k}\sum_{i=1}^{p}\sum_{l\ne i}sim(T^{f,neg}_{i,j}, T^{f,neg}_{l,j}).
\end{equation}

Overall, our NegPrompt objective consists of the above three losses. During training, we are able to obtain a diverse set of non-trivial prompts that effectively convey negative meanings relative to the ID class labels by minimizing the following overall loss:
\begin{equation}
    \mathcal{L}_{Negative Prompts} = \mathcal{L}_{NIS} + \beta * \mathcal{L}_{NPD} + \gamma * \mathcal{L}_{NND},
\end{equation}
where $\beta$ and $\gamma$ are hyper-parameters to balance the losses.
\vspace{-0.4cm}
\subsubsection{Open-Vocabulary Capability}
\vspace{-0.2cm}
Since the negative prompts we learn do not depend on specific class labels but are instead generic templates representing negative semantics of any given class labels, it is possible to utilize the generalization ability of CLIP to learn a set of transferable negative prompts. Specifically, instead of utilizing all of the ID classes $D_{train}^{in}$, NegPrompt may only employ its small subset $D_{train}^{in, sub}$ for training the negative prompts $\omega^{neg}$, where $D_{train}^{in, sub} = \{(x, y^{in}_{sub}) \mid   y^{in}_{sub} \subset y^{in}\}$. After obtaining the trained negative prompts, we combine them with the remaining ID class names, \ie, replace $c_i$ in $t^{neg}_{l,i}$ with the unseen ID class names, to obtain the corresponding negative prompts for the novel ID classes that are unseen during training. This approach, which achieves out-of-distribution detection by only exposing with a small portion of ID images, is unprecedented in prior research. We refer this as to be open-vocabulary OOD detection.

\subsubsection{Inference}
\vspace{-0.2cm}
During inference, we employ the MCM~\cite{ming2022delving} scoring approach for OOD detection, but with the addition of our negative prompts into the softmax function. Particularly, MCM uses the inverse of the maximum softmax score in Eq. \ref{eqn:softmax} as the OOD score. Our OOD scoring extends MCM and defines it as: $s(x)= \max (p(y=i|x))$, where $p(y=i|x)$ is defined in Eq. \ref{eq:softmax} that also includes the similarities of the test image to the negative prompts, in addition to the similarities to the positive prompts. The rationale behind this is that for ID images, they will be matched to one of the positive text features, leading to a higher $S^{f, pos}$ but lower $S^{f, neg}$, and thereby a higher maximum softmax score (\ie, a lower OOD score). Conversely, for OOD data, it will be matched to one of the negative text features, resulting in a lower maximum softmax score (\ie, a higher OOD score).

\vspace{-0.2cm}
\section{Experiment}
\subsection{Experimental Details}
\vspace{-0.2cm}

\begin{table}[t]
\centering
\scalebox{0.7}{
\begin{tabular}{@{}lll@{}}
\toprule
& \textbf{ID}          & \textbf{OOD}               \\ 
        \midrule
\textbf{Split-1}                & All dog classes               & Non-animal classes             \\
                        
\textbf{Split-2}                & Half of hunting dog classes       & Other 4-legged animal classes   \\
\textbf{Split-3}                & Mix of common classes  & Mix of common classes \\
 \bottomrule
\end{tabular}
}
\caption{Three ImageNet-1K splits for hard OOD detection.}
\label{tab:splits-information}
 \vspace{-0.6cm}
\end{table}

\noindent\textbf{Datasets.} For conventional OOD detection, we use a popular benchmark in which ImageNet-1K~\cite{deng2009imagenet} with 1,000 classes is used as the ID dataset, and the same OOD datasets as in ~\cite{ming2022delving} are used, including subsets of Texture~\cite{cimpoi2014describing}, iNaturalist~\cite{van2018inaturalist}, Places~\cite{zhou2017places} and SUN~\cite{xiao2010sun}. In addressing the more challenging OOD scenarios, we partitioned the ImageNet1k dataset into two segments: one segment of the data serves as the ID, while the other serves as OOD. As shown in Table \ref{tab:splits-information}, three different splits are derived, following from ~\cite{palechor2023large}. We further create another ImageNet split, Split-4, in which the first 100 classes are used as ID data and the subsequent 900 classes are used as OOD samples. Following CoOp and LoCoOp~\cite{zhou2022learning,miyai2023locoop}, during our training process, we utilized only few-shot training data for each category. Particularly, we only train the model with 16 images per ID class and without any exposure to OOD images. During testing, we employ the entire ID and OOD test set for evaluation.

\begin{table*}[t]
    \centering
    \scalebox{0.75}{
    \begin{tabular}{c cc  cc   cc  cc |cc}
        \hline
        \hline
        \specialrule{0em}{1pt}{1pt}
        \multirow{2}*{\textbf{Method}} & \multicolumn{2}{c}{\textbf{Texture}} &\multicolumn{2}{c}{\textbf{iNaturalist}} &\multicolumn{2}{c}{\textbf{Places}} & \multicolumn{2}{c|}{\textbf{SUN}} & \multicolumn{2}{c}{\textbf{Avg}}\\
        ~  &  AUC $\uparrow$  & FPR95 $\downarrow$ &  AUC $\uparrow$  & FPR95 $\downarrow$ &  AUC $\uparrow$  & FPR95 $\downarrow$ &  AUC $\uparrow$  & FPR95 $\downarrow$ &  AUC $\uparrow$  & FPR95 $\downarrow$\\
        \specialrule{0em}{1pt}{1pt}
        \hline
        \specialrule{0em}{1pt}{1pt}
        \textit{Zero-shot methods}& ~ & ~& ~& ~& ~& ~& ~& ~& ~& ~\\
        \textbf{MCM}~\cite{ming2022delving}$^\dag$ & 86.11 & 57.77 & 94.61 & 30.91 & 89.77 & 44.69 & 92.57 & 34.59 & 90.76 & 42.74\\
        \textbf{CLIPN}~\cite{wang2023clipn}$^\dag$ &  90.93 & 40.83 & 95.27 & 23.94 & 92.28 & 33.45 & 93.92 & 26.17 & 93.10 & 31.10\\
        \hline
        \textit{CLIP-based posthoc methods}& ~ & ~& ~& ~& ~& ~& ~& ~& ~& ~\\
        \textbf{MSP}~\cite{hendrycks2017a}$^\dag$ & 74.84 & 73.66 & 77.74 & 74.57 & 72.18 & 79.12 & 73.97 & 76.95 & 74.98 & 76.22 \\
        \textbf{MaxLogit}~\cite{hendrycks2019scaling}$^\dag$ & 88.63 & 48.72 & 88.03 & 60.88 & 87.45 & 55.54 & 91.16 & 44.83 & 88.82 & 52.49 \\
        \textbf{Energy}~\cite{liu2020energy}$^\dag$ & 88.22 & 50.39 & 87.18 & 64.98 & 87.33 & 57.40 & 91.17 & 46.42 & 88.48 & 54.80\\
        \textbf{ReAct}~\cite{sun2021react}$^\dag$ & 88.13 & 49.88 & 86.87 & 65.57 & 87.42 & 56.85 & 91.04 & 46.17 & 88.37 & 54.62\\
        \textbf{ODIN}~\cite{liang2018enhancing} $^\dag$& 87.85 & 51.67 & 94.65 & 30.22 & 85.54 & 55.06 & 87.17 & 54.04 & 88.80 & 47.75\\
        \hline
        \textit{Prompt learning methods} & ~ & ~& ~& ~& ~& ~& ~& ~& ~& ~\\
        \textbf{CoOp}~\cite{zhou2022learning} & 89.47 & 45.00 & 93.77 & 29.81 & 90.58 & 40.11 & 93.29 & 40.83 & 91.78 & 51.68\\
        \textbf{LoCoOp}~\cite{miyai2023locoop}$^\dag$ & 90.19 & 42.28 & 96.86 & 16.05 & 91.98 & 32.87 & 95.07 &    23.44 & 93.52 & 28.66\\
        \bf{NegPrompt} (Ours) & \textbf{91.60} & \textbf{35.21} &  \textbf{98.73} & \textbf{6.32} & \textbf{93.34} & \textbf{27.60} & \textbf{95.55} & \textbf{22.89} & \textbf{94.81} & \textbf{23.01} \\
        \hline\hline
        \textit{Open-vocabulary OOD detection} & ~ & ~& ~& ~& ~& ~& ~& ~& ~& ~\\
        \textbf{CoOp (10\%)} & 87.58 & 50.55 & 91.08 & 42.53 & 89.56 & 46.12 & 91.52 & 41.92 & 89.94 & 45.28\\
        \textbf{LoCoOp (10\%)} & 88.21 & 47.32  & 94.47 &  34.90 & 91.64 & 39.85 & 92.54 & 26.30 &  90.15 & 37.09\\
        \bf{NegPrompt} (Ours) (10\%) & 90.30 & 39.31 & 98.39 & 7.48 & 92.68 & 29.75 & 93.70 & 26.92 & 93.76 & 25.86\\
        \hline
        \hline
    \end{tabular}
    }
    \caption{Conventional OOD detection results. We trained using ImageNet1k as the ID and CLIP-B/16 as the CLIP backbone. The boldfaced results indicate the best performance. Results marked with $\dag$ are taken from ~\cite{wang2023clipn} and ~\cite{miyai2023locoop}. `METHOD' (10\%) in open-vocabulary OOD detection is to evaluate the performance of the `METHOD' when only images from 10\% ID classes are accessible during training.}
    \label{tab:main_results}
    \vspace{-0.2cm}
\end{table*}

\noindent\textbf{Implementation Details.} Following existing studies~\cite{wang2023clipn}, we use CLIP based on CLIP-B/16 which is pre-trained from OpenCLIP~\cite{ilharco_gabriel_2021_5143773}. NegPrompt is trained using 16-shot images of all ID classes under the normal OOD detection setting.
For open-vocabulary OOD detection, we train our model using only the images of the first 10\% classes from the ID dataset, withholding 90\% ID classes that only appear together with OOD data during inference.
We train a shared positive prompt and two shared negative prompts w.r.t. each training ID class.
The hyperparameters $\beta$ and $\gamma$ are set to 0.1 and 0.05, respectively (see Appendix A for detail). In the first stage, CoOp is trained for 100 epochs to obtain the positive prompts. In the second stage, the positive prompts are frozen, and our model is trained for 10 epochs to learn the negative prompts. For all experiments, we report the averaged results over three runs with different random seeds.

\noindent\textbf{Comparison Methods.} 
To substantiate the effectiveness of NegPrompts, we conduct an empirical analysis of three distinct categories of methodologies employed for OOD detection utilizing Vision-language models. These categories encompass zero-shot pretraining approaches,  the methods that combine the CLIP image encoder with classical approaches, and the methods grounded in prompt learning. In the context of zero-shot methods, we opted for the two recent methods, MCM~\cite{ming2022delving} and CLIPN~\cite{wang2023clipn}. MCM employs the original CLIP, utilizing the maximum softmax probability operation on the similarities for detection, and CLIPN involves an additional training phase during pretraining, specifically training a negative text encoder using large external data. For the second group of methods, we adapt previous logits-based methodologies to the use of the CLIP image encoder, including MSP~\cite{hendrycks2017a}, Energy~\cite{liu2020energy}, MaxLogit~\cite{hendrycks2019scaling}, ReAct~\cite{sun2021react} and ODIN~\cite{liang2018enhancing}, to serve as the CLIP-adapted methods. For the prompt learning methods, NegPrompt is compared with CoOp~\cite{zhou2022learning} and LoCoOp~\cite{miyai2023locoop}.

\noindent\textbf{Evaluation Metrics.}
Two OOD detection metrics are used. The first metric is the False Positive Rate at a 95\% True Negative Rate (FPR95), which denotes the rate of falsely identified OOD instances when the true negative rate is maintained at 95\%. The second metric is the Area Under the Receiver Operating Characteristic curve (AUROC), representing the measure of OOD ranking across various classification thresholds. We also check the classification accuracy of the ID data to evaluate how the OOD detectors affect the ID classification.

\begin{table*}[t]
    \centering
    \scalebox{0.75}{
    \begin{tabular}{c cc  cc   cc  cc |cc}
        \hline
        \hline
        \specialrule{0em}{1pt}{1pt}
        \multirow{2}*{\textbf{Method}} & \multicolumn{2}{c}{\textbf{Split-1}} &\multicolumn{2}{c}{\textbf{Split-2}} &\multicolumn{2}{c}{\textbf{Split-3}} & \multicolumn{2}{c|}{\textbf{Split-4}} & \multicolumn{2}{c}{\textbf{Avg}}\\
        ~  &  AUC $\uparrow$  & FPR95 $\downarrow$ &  AUC $\uparrow$  & FPR95 $\downarrow$ &  AUC $\uparrow$  & FPR95 $\downarrow$ &  AUC $\uparrow$  & FPR95 $\downarrow$ &  AUC $\uparrow$  & FPR95 $\downarrow$\\
        \specialrule{0em}{1pt}{1pt}
        \hline
        \specialrule{0em}{1pt}{1pt}
        \textit{Zero-shot methods}& ~ & ~& ~& ~& ~& ~& ~& ~& ~& ~\\
        \textbf{MCM}& 97.93 & 9.17 & 88.10 & 56.40 & 90.34 & 33.05 & 98.72 & 4.73 & 93.77 & 25.83\\
        \textbf{CLIPN}& 99.38 & 2.07 & 97.77 & 10.55 & 90.03 & 36.85 & 98.83 & 4.68 & 96.50 & 13.53\\
        \hline
        \textit{CLIP-based posthoc methods}& ~ & ~& ~& ~& ~& ~& ~& ~& ~& ~\\
        \textbf{MSP} & 77.85 & 63.60 & 68.73 & 83.63 & 79.10 & 70.55 & 82.40 & 65.52 & 77.02 & 70.83 \\
        \textbf{MaxLogit} & 99.87 & 0.49 & 98.06 & 8.69 & 90.96 & 34.34 & 99.35 & 2.66 & 97.06 & 11.55 \\
        \textbf{Energy} & \textbf{99.88} & \textbf{0.46} & 98.18 & 8.40 & 90.65 & 35.02 & 99.36 & 2.83 & 97.02 & 11.68\\
        \textbf{ReAct} & 99.34 & 0.72 & 97.91 & 9.33 & 90.72 & 35.65 & 99.12 & 2.94 & 96.77 & 12.16\\
        \textbf{ODIN}& 98.78 & 1.12 & 98.23 & 8.18 & 89.92 & 37.20 & 98.76 & 13.20 & 96.42 & 14.92\\
        \hline
        \textit{Prompt learning methods} & ~ & ~& ~& ~& ~& ~& ~& ~& ~& ~\\
        \textbf{CoOp} & 98.53 & 6.78 & 88.25 & 50.76 & 90.64 & 33.89 & 98.54 & 5.11 & 93.99 & 24.14\\
        \textbf{LoCoOp} & 98.64 & 6.29 & 84.63 & 61.09 & 91.30 & 27.79 & 98.83 & 41.44 & 93.35 & 34.15\\
        \bf{NegPrompt} (Ours) & 99.85 & 0.62 & \textbf{98.54} & \textbf{7.60} & \textbf{93.89} & \textbf{22.89} & \textbf{99.57} & \textbf{1.60} & \textbf{97.96} & \textbf{8.18} \\
        \hline\hline
        \textit{Open-vocabulary OOD detection} & ~ & ~& ~& ~& ~& ~& ~& ~& ~& ~\\
        \textbf{CoOp (10\%)} & 97.97 & 12.217 & 80.11 & 74.62 & 87.92 & 46.00 & 96.59 & 16.60 & 90.65 & 37.36\\
        \textbf{LoCoOp (10\%)} & 98.00 & 9.23 & 87.02 & 52.18 & 80.51 & 59.93 & 82.41 & 48.72 & 86.99 & 42.52\\
        \bf{NegPrompt} (Ours) (10\%) & 99.66 & 1.36 & 96.30 & 19.89 & 91.75 & 26.92 & 98.14 & 5.24 & 96.46 & 13.36\\

        \hline
        \hline
    \end{tabular}
    }
    \caption{Hard OOD detection results. We use the same notations here as those used in Table 2. }
    \label{tab:open_set results}
    \vspace{-0.2cm}
\end{table*}
\subsection{Comparison to State-of-the-art Models}
\vspace{-0.2cm}
\noindent\textbf{Conventional OOD Detection.}
The results of conventional OOD detection are reported in Table \ref{tab:main_results}. It is clear that our proposed NegPrompt achieves consistently superior performance in both individual OOD datasets and the averaged results. When compared with the zero-shot methods, on average, our approach surpasses the best competing method CLIPN by more than 1.5\% in AUC and around 8\% in FPR95, despite the fact that CLIPN requires the use of an additional large external dataset to train an additional negative text encoder. In other words, although NegPrompt is significantly more lightweight than CLIPN in model size, it can substantially and consistently outperform CLIPN in both metrics across all OOD datasets. The adapted post-hoc methods generally do not leverage the CLIP's capabilities well and thus perform less effectively.

NegPrompt also substantially surpasses both prompt learning-based methods, reducing the FPR95 by about 28\% (CoOp) and 5\% (LoCoOp). This indicates that the learned negative prompts provide informed knowledge about OOD data, which is lacking in the competing methods, helping largely reduce detection errors.

\noindent\textbf{Hard OOD Detection.}
Hard OOD detection presents unique challenges as the OOD samples often exhibit some similar features as the ID samples.
The results on the four hard OOD datasets derived from ImageNet-1K are shown in Table \ref{tab:open_set results}.
Similar empirical observations can be derived. Our method NegPropmt is consistently the best performer in the average performance, showcasing its general effectiveness across different dataset splits. The superiority of NegPrompt over the zero-shot and prompt learning-based methods is similar to that in Table \ref{tab:main_results}. Although it is slightly less effective than Energy under the Split-1 setting, it outperforms Energy in all other metrics, achieving maximally over 3\% AUC and 12\% FPR95 improvement among the performance on the other OOD datasets. 

Note that the results in Table \ref{tab:open_set results} are generally more promising than that in Table \ref{tab:main_results}. This is mainly because the OOD detection difficulty in all four ImageNet-1K splits is largely reduced since the number of their ID classes is significantly less than that in the full ImageNet-1K data.

\noindent\textbf{Open-Vocabulary OOD Detection.} 
The open-vocabulary OOD detection results are reported in both Tables \ref{tab:main_results} and \ref{tab:open_set results}. Impressively, even when training on only 10\% ID classes, our approach can still perform better than the competing methods using the full ID classes, \eg, the average AUC and FPR95 in Table \ref{tab:main_results}. 
In this open-vocabulary setting, in general, both LoCoOp and CoOp exhibit a much larger performance decline than NegPrompt, especially on the results in Table \ref{tab:open_set results}, in which CoOp has over 3\% AUC drop and LoCoOp has over 6\% AUC drop while our method has only about 1.5\% AUC drop. These results demonstrate that the negative prompts in NegPrompt have much better transferability than those in the two competing methods.

\subsubsection{Classification Accuracy on ID Data}
\vspace{-0.2cm}
We also evaluate the classification accuracy on the ID data when using NegPrompt for OOD detection, with CoOp, LoCoOp, MCM and CLIPN as the baselines. The classification accuracy results on the full ImageNet-1K test data are shown in Table \ref{tab:acc_comparison}. When using the full ImageNet-1K training ID class data, our method NegPrompt can maintain the same classification accuracy as CoOp. 

Our accuracy is slightly compromised when using only 10\% ID classes in our training. On the other hand, the OOD detection in LoCoOp compromises the ID classification accuracy, dropped from 72.1\% in CoOp to 71.7\%. This may be attributed to its focus on the localized regions within the background rather than the primary object of interest, resulting in the missing of some discriminative features for ID data classification. CLIPN and MCM, being zero-shot methods, have not been exposed to the target ID data, leading to a much lower accuracy than the other methods. 

\begin{table}[t]
    \centering
    \scalebox{0.80}{
    \begin{tabular}{lc}
        \toprule
        \textbf{Method} & \textbf{Top-1 Accuracy} \\
        \midrule
        CoOp & 72.1 \\
        LoCoOp$^\dag$ & 71.7 \\
        CLIPN \& MCM & 67.0\\
        \textbf{NegPrompt(Ours)(10\%)} & 71.9\\
        \textbf{NegPrompt(Ours)(Full)} & 72.1\\
        \bottomrule
    \end{tabular}
    }
    \caption{Top-1 Accuracy.
    Results with $\dag$ are taken from ~\cite{miyai2023locoop}.}
    \label{tab:acc_comparison}
    \vspace{-0.3cm}
\end{table}

\begin{figure}
  \centering
    \centering
    \includegraphics[width=0.46\linewidth]{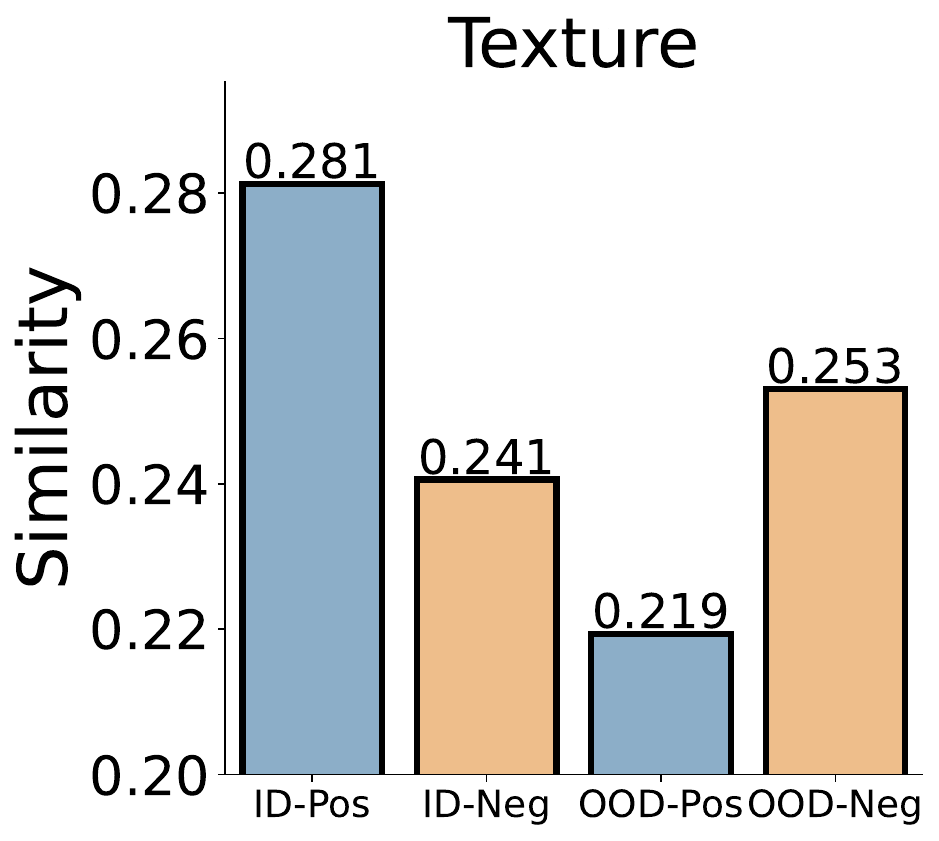}
    \hfill
    \includegraphics[width=0.46\linewidth]{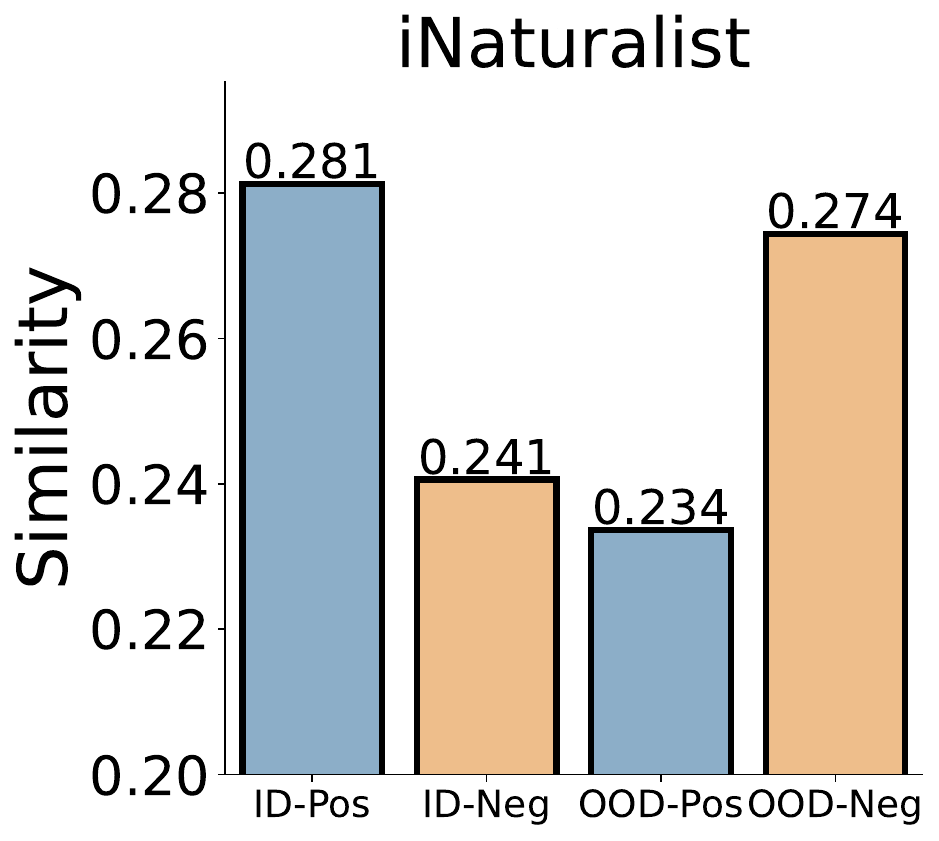}
    \includegraphics[width=0.46\linewidth]{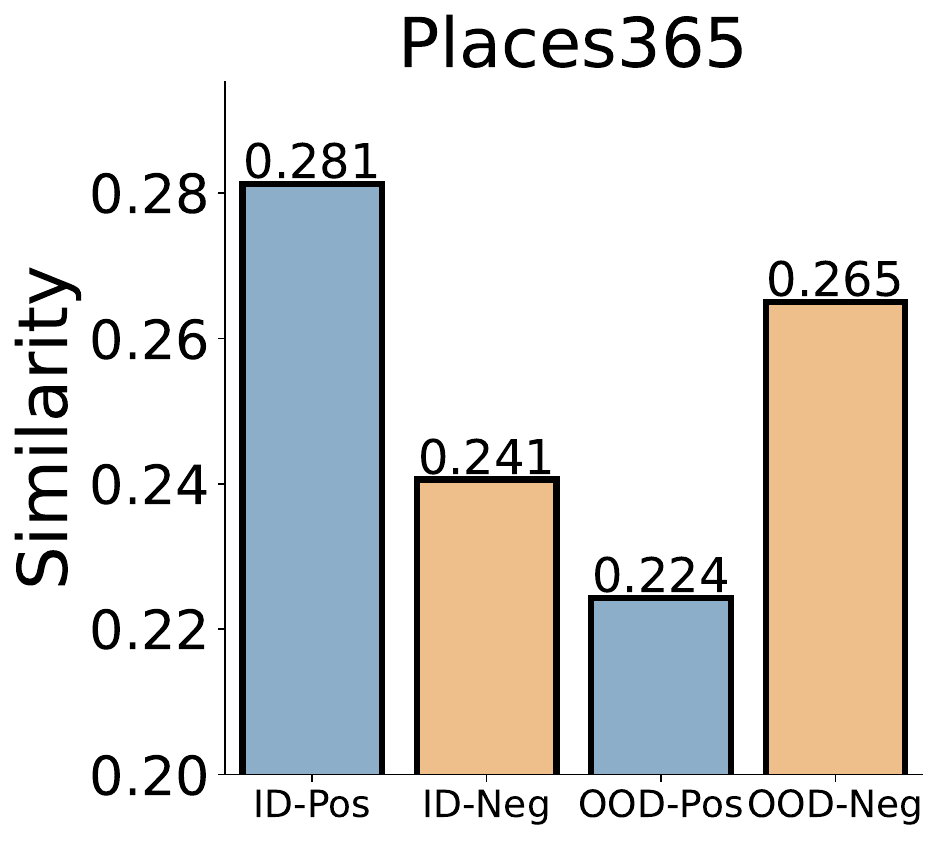}
    \hfill
    \includegraphics[width=0.46\linewidth]{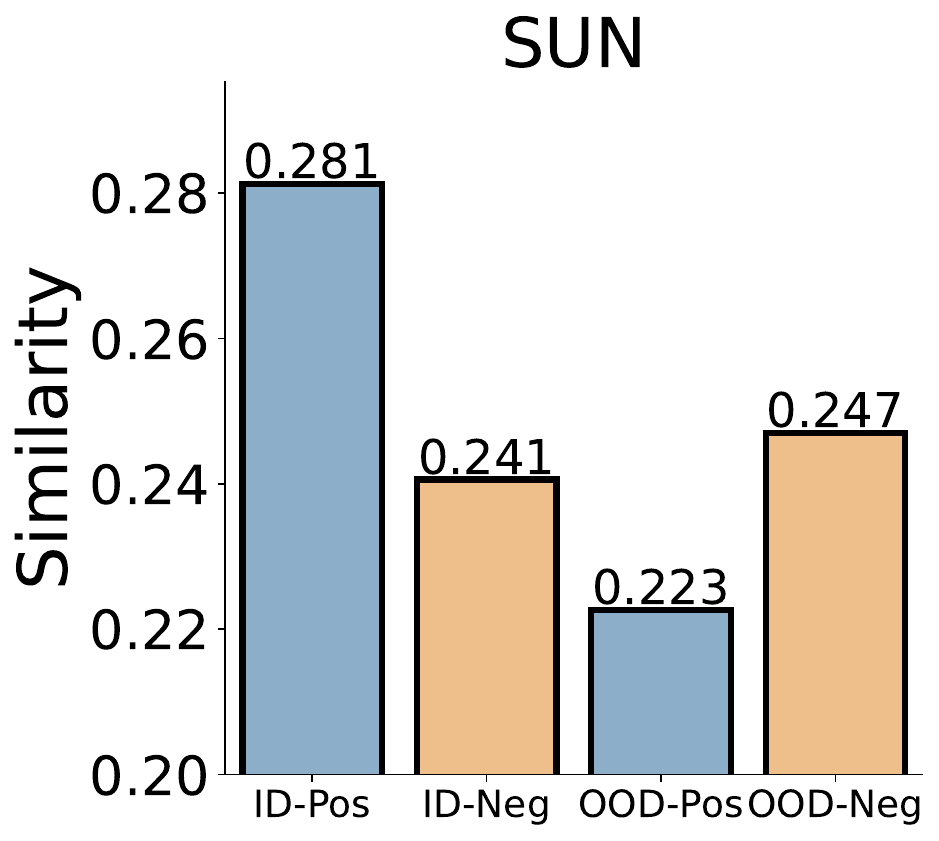}
  \caption{Similarity of ID/OOD and Positive/Negative Prompts.}
  \label{fig:similarity}
\end{figure}
\begin{figure}[ht]

  \centering
    \includegraphics[width=0.85\linewidth]{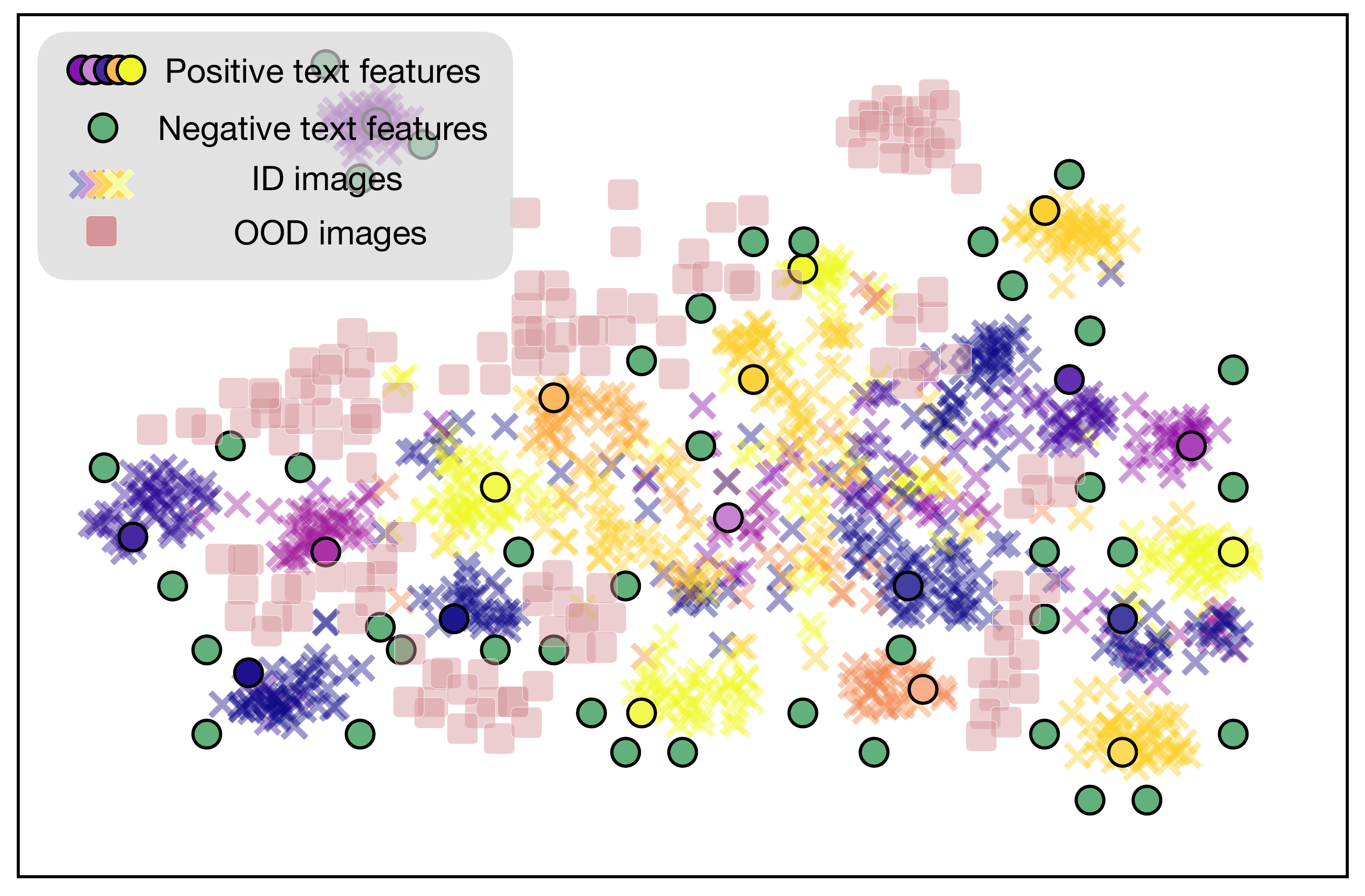}
  \caption{T-SNE visualization of NegPrompt, utilizing a subset of ImageNet - TinyImageNet as the dataset. }
  \label{fig:t-sne}

  \vspace{-0.2cm}
\end{figure}

\begin{table*}[ht]
    \centering
    \scalebox{0.75}{
    \begin{tabular}{c cc  cc   cc  cc |ccc}
        \hline
        \hline
        \specialrule{0em}{1pt}{1pt}
        \multirow{2}*{\textbf{Ablation Study}} & \multicolumn{2}{c}{\textbf{Texture}} &\multicolumn{2}{c}{\textbf{iNaturalist}} &\multicolumn{2}{c}{\textbf{Places}} & \multicolumn{2}{c|}{\textbf{SUN}} & \multicolumn{3}{c}{\textbf{Avg}}\\
        ~  &  AUC $\uparrow$  & FPR95 $\downarrow$ &  AUC $\uparrow$  & FPR95 $\downarrow$ &  AUC $\uparrow$  & FPR95 $\downarrow$ &  AUC $\uparrow$  & FPR95 $\downarrow$ &  AUC $\uparrow$  & FPR95 $\downarrow$ & ACC $\uparrow$\\
        \specialrule{0em}{1pt}{1pt}
        \hline
        \specialrule{0em}{1pt}{1pt}
        \textit{ Backbones}& ~ & ~& ~& ~& ~& ~& ~& ~& ~& ~\\
        \textbf{ResNet-50} & 79.44 & 72.73 & 90.97  & 44.10 & 84.35 & 61.37 & 88.04 & 50.55 & 85.70 & 57.19 & 68.2\\
        \textbf{ResNet-101} & 82.97 & 70.83 & 93.96 & 31.3 & 86.41 & 51.66 & 88.81 & 47.61 & 88.04 & 50.35 & 70.3\\
        \textbf{ViT-B-32} & \textbf{90.43} & \textbf{38.79} & 97.40 & 12.64 & \textbf{92.83} & 32.79 & 92.82 & \textbf{26.03} & 93.37 & 27.56 & \textbf{72.6}\\
        \hline
        \textit{\# Negative Prompts}& ~ & ~& ~& ~& ~& ~& ~& ~& ~& ~\\
        \textbf{1} & 90.04 & 40.67 & 98.24 & 9.23 & 90.37 & 32.62 & 92.26 & 27.33 & 92.73 & 27.46 & 72.1\\
        \hline
        \textit{Training Process} & ~ & ~& ~& ~& ~& ~& ~& ~& ~& ~\\
        \textbf{One-stage} & 80.25 & 72.87 & 77.02 & 95.74 & 82.53 & 95.94 & 81.13 & 95.72 & 80.23 & 90.07 & 62.8\\
        \specialrule{0em}{1pt}{1pt}
        \hline
        \specialrule{0em}{1pt}{1pt}
        \bf{ViT-B-16 \& 2 \& Two-stage}& 90.30 & 39.31 & \textbf{98.39} & \textbf{7.48} & 92.68 & \textbf{29.75} & \textbf{93.70} & 26.92 & \textbf{93.76} & \textbf{25.86} & 72.1\\
        \specialrule{0em}{1pt}{1pt}
        \hline
        \hline
    \end{tabular}
    }
    \caption{Results of our ablation experiments. }
    \label{tab:ablation}
    \vspace{-0.3cm}
\end{table*}

\vspace{-0.2cm}
\subsection{Analysis of NegPrompt}
\subsubsection{Why Does NegPrompt Work?}
\vspace{-0.2cm}
To better understand the effectiveness of NegPrompt, we summarize the average maximum similarity between ID/OOD images and positive/negative prompts, as shown in Fig. \ref{fig:similarity}. Across all four OOD datasets, ID images have the highest similarity with positive prompts, and OOD images with negative prompts. This suggests positive prompts are closer to ID images and negative prompts to OOD images in latent space, ensuring OOD images receive lower softmax scores (\ie, higher OOD scores) than ID images in Eq. \ref{eq:softmax}.

Using TinyImageNet~\cite{le2015tiny}, a subset of ImageNet, we further visualized the learned negative text features with its test data~\cite{esmaeilpour2022zero}. Learning three negative prompts per ID class, the results in Fig. \ref{fig:t-sne} demonstrate that positive text features from positive prompts align closely with ID images in latent space. Conversely, the negative text features encoded from negative prompts lie outside of the ID data, with OOD images interspersed among them. This suggests negative text features effectively act as a fence aligning much better to the OOD images than the ID images, well supporting the OOD detection while preserving the ID classification accuracy.
\vspace{-0.2cm}
\subsubsection{Ablation Study}
\vspace{-0.3cm}

Table \ref{tab:ablation} shows our ablation study results on the backbone, the number of negative prompts, and the training process.

\noindent \textbf{Backbones.} We experiment with diverse CLIP backbones. The results reveal that for CNN-based backbones like ResNet50 and ResNet101, the OOF detection performance is not as proficient as the ViT-based backbones. Regarding ViT-B-32 and ViT-B-16, their performance is found to be uneven. Overall, the performance of OOD detection tends to increase with more advanced backbones.

\noindent \textbf{The Number of Negative Prompts.} The number of negative prompts also influences the results. It is observed that the OOD detection performance improves when increasing the number of negative prompts from one to two. Therefore, it is suggested to further increase the number of negative prompts for better detection accuracy. However, note that with an increase in the number of negative prompts, the computational cost also rises rapidly. A good balance between computational cost and detection accuracy is needed when determining the number of negative prompts.

\noindent \textbf{Training Process.} The training process is also important. As discussed before, due to the necessity of anchoring the positive prompts, a two-stage training process is used in our model. This process involves training the positive prompts in the first stage and freezing them before proceeding to train the negative prompts in the second stage. When simultaneously training both positive and negative prompts in a unified step, the model's ability to effectively learn positive prompts is significantly undermined, and consequently we obtain unstable negative prompts, leading to the largely decreased OOD detection performance.

\section{Conclusion}

We present NegPrompt, a novel approach for prompt learning-based OOD detection. It utilizes VLMs to learn a small set of negative prompts for conveying negative semantics relative to ID classes. Our empirical results reveal that NegPrompt 1) achieves superior OOD detection performance compared to the SOTA models across various OOD datasets in both conventional and hard OOD detection scenarios, and 2) learns transfer negative prompts that enable  excellent open-vocabulary OOD detection performance.

\section{Acknowledgement}
This work was supported by the National Natural Science Foundation of China 62276016, 62372029.
{
    \small
    \bibliographystyle{ieeenat_fullname}

    \bibliography{camera_ready}

\begin{thebibliography}{57}
\providecommand{\natexlab}[1]{#1}
\providecommand{\url}[1]{\texttt{#1}}
\expandafter\ifx\csname urlstyle\endcsname\relax
  \providecommand{\doi}[1]{doi: #1}\else
  \providecommand{\doi}{doi: \begingroup \urlstyle{rm}\Url}\fi

\bibitem[Alayrac et~al.(2022)Alayrac, Donahue, Luc, Miech, Barr, Hasson, Lenc, Mensch, Millican, Reynolds, et~al.]{alayrac2022flamingo}
Jean-Baptiste Alayrac, Jeff Donahue, Pauline Luc, Antoine Miech, Iain Barr, Yana Hasson, Karel Lenc, Arthur Mensch, Katherine Millican, Malcolm Reynolds, et~al.
\newblock Flamingo: a visual language model for few-shot learning.
\newblock \emph{Advances in Neural Information Processing Systems}, 35:\penalty0 23716--23736, 2022.

\bibitem[Cimpoi et~al.(2014)Cimpoi, Maji, Kokkinos, Mohamed, and Vedaldi]{cimpoi2014describing}
Mircea Cimpoi, Subhransu Maji, Iasonas Kokkinos, Sammy Mohamed, and Andrea Vedaldi.
\newblock Describing textures in the wild.
\newblock In \emph{Proceedings of the IEEE conference on computer vision and pattern recognition}, pages 3606--3613, 2014.

\bibitem[Deng et~al.(2009)Deng, Dong, Socher, Li, Li, and Fei-Fei]{deng2009imagenet}
Jia Deng, Wei Dong, Richard Socher, Li-Jia Li, Kai Li, and Li Fei-Fei.
\newblock Imagenet: A large-scale hierarchical image database.
\newblock In \emph{2009 IEEE conference on computer vision and pattern recognition}, pages 248--255. Ieee, 2009.

\bibitem[Devlin et~al.(2018)Devlin, Chang, Lee, and Toutanova]{devlin2018bert}
Jacob Devlin, Ming-Wei Chang, Kenton Lee, and Kristina Toutanova.
\newblock Bert: Pre-training of deep bidirectional transformers for language understanding.
\newblock \emph{arXiv preprint arXiv:1810.04805}, 2018.

\bibitem[Dosovitskiy et~al.(2020)Dosovitskiy, Beyer, Kolesnikov, Weissenborn, Zhai, Unterthiner, Dehghani, Minderer, Heigold, Gelly, et~al.]{dosovitskiy2020image}
Alexey Dosovitskiy, Lucas Beyer, Alexander Kolesnikov, Dirk Weissenborn, Xiaohua Zhai, Thomas Unterthiner, Mostafa Dehghani, Matthias Minderer, Georg Heigold, Sylvain Gelly, et~al.
\newblock An image is worth 16x16 words: Transformers for image recognition at scale.
\newblock \emph{arXiv preprint arXiv:2010.11929}, 2020.

\bibitem[Esmaeilpour et~al.(2022)Esmaeilpour, Liu, Robertson, and Shu]{esmaeilpour2022zero}
Sepideh Esmaeilpour, Bing Liu, Eric Robertson, and Lei Shu.
\newblock Zero-shot out-of-distribution detection based on the pre-trained model clip.
\newblock In \emph{Proceedings of the AAAI conference on artificial intelligence}, pages 6568--6576, 2022.

\bibitem[Fang et~al.(2023)Fang, Pang, and Bai]{fang2023simple}
Ruohuan Fang, Guansong Pang, and Xiao Bai.
\newblock Simple image-level classification improves open-vocabulary object detection.
\newblock \emph{arXiv preprint arXiv:2312.10439}, 2023.

\bibitem[Gao et~al.(2023)Gao, Geng, Zhang, Ma, Fang, Zhang, Li, and Qiao]{gao2023clip}
Peng Gao, Shijie Geng, Renrui Zhang, Teli Ma, Rongyao Fang, Yongfeng Zhang, Hongsheng Li, and Yu Qiao.
\newblock Clip-adapter: Better vision-language models with feature adapters.
\newblock \emph{International Journal of Computer Vision}, pages 1--15, 2023.

\bibitem[He et~al.(2016)He, Zhang, Ren, and Sun]{he2016deep}
Kaiming He, Xiangyu Zhang, Shaoqing Ren, and Jian Sun.
\newblock Deep residual learning for image recognition.
\newblock In \emph{Proceedings of the IEEE conference on computer vision and pattern recognition}, pages 770--778, 2016.

\bibitem[Hendrycks and Gimpel(2017)]{hendrycks2017a}
Dan Hendrycks and Kevin Gimpel.
\newblock A baseline for detecting misclassified and out-of-distribution examples in neural networks.
\newblock In \emph{International Conference on Learning Representations}, 2017.

\bibitem[Hendrycks et~al.(2018)Hendrycks, Mazeika, and Dietterich]{hendrycks2018deep}
Dan Hendrycks, Mantas Mazeika, and Thomas Dietterich.
\newblock Deep anomaly detection with outlier exposure.
\newblock \emph{arXiv preprint arXiv:1812.04606}, 2018.

\bibitem[Hendrycks et~al.(2019)Hendrycks, Basart, Mazeika, Zou, Kwon, Mostajabi, Steinhardt, and Song]{hendrycks2019scaling}
Dan Hendrycks, Steven Basart, Mantas Mazeika, Andy Zou, Joe Kwon, Mohammadreza Mostajabi, Jacob Steinhardt, and Dawn Song.
\newblock Scaling out-of-distribution detection for real-world settings.
\newblock \emph{arXiv preprint arXiv:1911.11132}, 2019.

\bibitem[Hu et~al.(2023)Hu, Sun, Sclaroff, and Saenko]{hu2023dualcoop++}
Ping Hu, Ximeng Sun, Stan Sclaroff, and Kate Saenko.
\newblock Dualcoop++: Fast and effective adaptation to multi-label recognition with limited annotations.
\newblock \emph{IEEE Transactions on Pattern Analysis and Machine Intelligence}, 2023.

\bibitem[Huang and Li(2021)]{huang2021mos}
Rui Huang and Yixuan Li.
\newblock Mos: Towards scaling out-of-distribution detection for large semantic space.
\newblock In \emph{Proceedings of the IEEE/CVF Conference on Computer Vision and Pattern Recognition}, pages 8710--8719, 2021.

\bibitem[Huang et~al.(2021)Huang, Geng, and Li]{huang2021importance}
Rui Huang, Andrew Geng, and Yixuan Li.
\newblock On the importance of gradients for detecting distributional shifts in the wild.
\newblock \emph{Advances in Neural Information Processing Systems}, 34:\penalty0 677--689, 2021.

\bibitem[Ilharco et~al.(2021)Ilharco, Wortsman, Wightman, Gordon, Carlini, Taori, Dave, Shankar, Namkoong, Miller, Hajishirzi, Farhadi, and Schmidt]{ilharco_gabriel_2021_5143773}
Gabriel Ilharco, Mitchell Wortsman, Ross Wightman, Cade Gordon, Nicholas Carlini, Rohan Taori, Achal Dave, Vaishaal Shankar, Hongseok Namkoong, John Miller, Hannaneh Hajishirzi, Ali Farhadi, and Ludwig Schmidt.
\newblock Openclip, 2021.

\bibitem[Jia et~al.(2021)Jia, Yang, Xia, Chen, Parekh, Pham, Le, Sung, Li, and Duerig]{jia2021scaling}
Chao Jia, Yinfei Yang, Ye Xia, Yi-Ting Chen, Zarana Parekh, Hieu Pham, Quoc Le, Yun-Hsuan Sung, Zhen Li, and Tom Duerig.
\newblock Scaling up visual and vision-language representation learning with noisy text supervision.
\newblock In \emph{International conference on machine learning}, pages 4904--4916, 2021.

\bibitem[Jia et~al.(2022)Jia, Tang, Chen, Cardie, Belongie, Hariharan, and Lim]{jia2022visual}
Menglin Jia, Luming Tang, Bor-Chun Chen, Claire Cardie, Serge Belongie, Bharath Hariharan, and Ser-Nam Lim.
\newblock Visual prompt tuning.
\newblock In \emph{European Conference on Computer Vision}, pages 709--727. Springer, 2022.

\bibitem[Jiang et~al.(2020)Jiang, Xu, Araki, and Neubig]{jiang2020can}
Zhengbao Jiang, Frank~F Xu, Jun Araki, and Graham Neubig.
\newblock How can we know what language models know?
\newblock \emph{Transactions of the Association for Computational Linguistics}, 8:\penalty0 423--438, 2020.

\bibitem[Khattak et~al.(2023)Khattak, Rasheed, Maaz, Khan, and Khan]{khattak2023maple}
Muhammad~Uzair Khattak, Hanoona Rasheed, Muhammad Maaz, Salman Khan, and Fahad~Shahbaz Khan.
\newblock Maple: Multi-modal prompt learning.
\newblock In \emph{Proceedings of the IEEE/CVF Conference on Computer Vision and Pattern Recognition}, pages 19113--19122, 2023.

\bibitem[Khosla et~al.(2020)Khosla, Teterwak, Wang, Sarna, Tian, Isola, Maschinot, Liu, and Krishnan]{khosla2020supervised}
Prannay Khosla, Piotr Teterwak, Chen Wang, Aaron Sarna, Yonglong Tian, Phillip Isola, Aaron Maschinot, Ce Liu, and Dilip Krishnan.
\newblock Supervised contrastive learning.
\newblock \emph{Advances in neural information processing systems}, 33:\penalty0 18661--18673, 2020.

\bibitem[Kong and Ramanan(2021)]{kong2021opengan}
Shu Kong and Deva Ramanan.
\newblock Opengan: Open-set recognition via open data generation.
\newblock In \emph{Proceedings of the IEEE/CVF International Conference on Computer Vision}, pages 813--822, 2021.

\bibitem[Le and Yang(2015)]{le2015tiny}
Ya Le and Xuan Yang.
\newblock Tiny imagenet visual recognition challenge.
\newblock \emph{CS 231N}, 7\penalty0 (7):\penalty0 3, 2015.

\bibitem[Li et~al.(2022)Li, Li, Xiong, and Hoi]{li2022blip}
Junnan Li, Dongxu Li, Caiming Xiong, and Steven Hoi.
\newblock Blip: Bootstrapping language-image pre-training for unified vision-language understanding and generation.
\newblock In \emph{International Conference on Machine Learning}, pages 12888--12900. PMLR, 2022.

\bibitem[Li et~al.(2024)Li, Pang, Bai, Zheng, Zhou, and Ning]{LI2024110258}
Tianqi Li, Guansong Pang, Xiao Bai, Jin Zheng, Lei Zhou, and Xin Ning.
\newblock Learning adversarial semantic embeddings for zero-shot recognition in open worlds.
\newblock \emph{Pattern Recognition}, 149:\penalty0 110258, 2024.

\bibitem[Liang et~al.(2018)Liang, Li, and Srikant]{liang2018enhancing}
Shiyu Liang, Yixuan Li, and R. Srikant.
\newblock Enhancing the reliability of out-of-distribution image detection in neural networks.
\newblock In \emph{International Conference on Learning Representations}, 2018.

\bibitem[Liu et~al.(2020)Liu, Wang, Owens, and Li]{liu2020energy}
Weitang Liu, Xiaoyun Wang, John Owens, and Yixuan Li.
\newblock Energy-based out-of-distribution detection.
\newblock \emph{Advances in neural information processing systems}, 33:\penalty0 21464--21475, 2020.

\bibitem[Liu et~al.(2023)Liu, Ding, Tian, Pang, Belagiannis, Reid, and Carneiro]{liu2023residual}
Yuyuan Liu, Choubo Ding, Yu Tian, Guansong Pang, Vasileios Belagiannis, Ian Reid, and Gustavo Carneiro.
\newblock Residual pattern learning for pixel-wise out-of-distribution detection in semantic segmentation.
\newblock In \emph{Proceedings of the IEEE/CVF International Conference on Computer Vision}, pages 1151--1161, 2023.

\bibitem[Liu et~al.(2021)Liu, Lin, Cao, Hu, Wei, Zhang, Lin, and Guo]{liu2021swin}
Ze Liu, Yutong Lin, Yue Cao, Han Hu, Yixuan Wei, Zheng Zhang, Stephen Lin, and Baining Guo.
\newblock Swin transformer: Hierarchical vision transformer using shifted windows.
\newblock In \emph{Proceedings of the IEEE/CVF international conference on computer vision}, pages 10012--10022, 2021.

\bibitem[Miao et~al.(2023)Miao, Pang, Li, Bai, and Zheng]{miao2023out}
Wenjun Miao, Guansong Pang, Tianqi Li, Xiao Bai, and Jin Zheng.
\newblock Out-of-distribution detection in long-tailed recognition with calibrated outlier class learning.
\newblock \emph{arXiv preprint arXiv:2312.10686}, 2023.

\bibitem[Minderer et~al.(2022)Minderer, Gritsenko, Stone, Neumann, Weissenborn, Dosovitskiy, Mahendran, Arnab, Dehghani, Shen, et~al.]{minderer2022simple}
Matthias Minderer, Alexey Gritsenko, Austin Stone, Maxim Neumann, Dirk Weissenborn, Alexey Dosovitskiy, Aravindh Mahendran, Anurag Arnab, Mostafa Dehghani, Zhuoran Shen, et~al.
\newblock Simple open-vocabulary object detection.
\newblock In \emph{European Conference on Computer Vision}, pages 728--755. Springer, 2022.

\bibitem[Ming et~al.(2022)Ming, Cai, Gu, Sun, Li, and Li]{ming2022delving}
Yifei Ming, Ziyang Cai, Jiuxiang Gu, Yiyou Sun, Wei Li, and Yixuan Li.
\newblock Delving into out-of-distribution detection with vision-language representations.
\newblock \emph{Advances in Neural Information Processing Systems}, 35:\penalty0 35087--35102, 2022.

\bibitem[Miyai et~al.(2023)Miyai, Yu, Irie, and Aizawa]{miyai2023locoop}
Atsuyuki Miyai, Qing Yu, Go Irie, and Kiyoharu Aizawa.
\newblock Locoop: Few-shot out-of-distribution detection via prompt learning.
\newblock In \emph{Thirty-Seventh Conference on Neural Information Processing Systems}, 2023.

\bibitem[Nguyen et~al.(2015)Nguyen, Yosinski, and Clune]{nguyen2015deep}
Anh Nguyen, Jason Yosinski, and Jeff Clune.
\newblock Deep neural networks are easily fooled: High confidence predictions for unrecognizable images.
\newblock In \emph{CVPR}, pages 427--436, 2015.

\bibitem[Palechor et~al.(2023)Palechor, Bhoumik, and G{\"u}nther]{palechor2023large}
Andres Palechor, Annesha Bhoumik, and Manuel G{\"u}nther.
\newblock Large-scale open-set classification protocols for imagenet.
\newblock In \emph{Proceedings of the IEEE/CVF Winter Conference on Applications of Computer Vision}, pages 42--51, 2023.

\bibitem[Radford et~al.(2018)Radford, Narasimhan, Salimans, Sutskever, et~al.]{radford2018improving}
Alec Radford, Karthik Narasimhan, Tim Salimans, Ilya Sutskever, et~al.
\newblock Improving language understanding by generative pre-training.
\newblock 2018.

\bibitem[Radford et~al.(2021)Radford, Kim, Hallacy, Ramesh, Goh, Agarwal, Sastry, Askell, Mishkin, Clark, Krueger, and Sutskever]{radford2021learning}
Alec Radford, Jong~Wook Kim, Chris Hallacy, Aditya Ramesh, Gabriel Goh, Sandhini Agarwal, Girish Sastry, Amanda Askell, Pamela Mishkin, Jack Clark, Gretchen Krueger, and Ilya Sutskever.
\newblock Learning transferable visual models from natural language supervision, 2021.

\bibitem[Schuhmann et~al.(2022{\natexlab{a}})Schuhmann, Beaumont, Vencu, Gordon, Wightman, Cherti, Coombes, Katta, Mullis, Wortsman, et~al.]{schuhmann2022laion}
Christoph Schuhmann, Romain Beaumont, Richard Vencu, Cade Gordon, Ross Wightman, Mehdi Cherti, Theo Coombes, Aarush Katta, Clayton Mullis, Mitchell Wortsman, et~al.
\newblock Laion-5b: An open large-scale dataset for training next generation image-text models.
\newblock \emph{Advances in Neural Information Processing Systems}, 35:\penalty0 25278--25294, 2022{\natexlab{a}}.

\bibitem[Schuhmann et~al.(2022{\natexlab{b}})Schuhmann, Beaumont, Vencu, Gordon, Wightman, Cherti, Coombes, Katta, Mullis, Wortsman, Schramowski, Kundurthy, Crowson, Schmidt, Kaczmarczyk, and Jitsev]{schuhmann2022laionb}
Christoph Schuhmann, Romain Beaumont, Richard Vencu, Cade~W Gordon, Ross Wightman, Mehdi Cherti, Theo Coombes, Aarush Katta, Clayton Mullis, Mitchell Wortsman, Patrick Schramowski, Srivatsa~R Kundurthy, Katherine Crowson, Ludwig Schmidt, Robert Kaczmarczyk, and Jenia Jitsev.
\newblock {LAION}-5b: An open large-scale dataset for training next generation image-text models.
\newblock In \emph{Thirty-sixth Conference on Neural Information Processing Systems Datasets and Benchmarks Track}, 2022{\natexlab{b}}.

\bibitem[Shin et~al.(2020)Shin, Razeghi, Logan~IV, Wallace, and Singh]{shin2020autoprompt}
Taylor Shin, Yasaman Razeghi, Robert~L Logan~IV, Eric Wallace, and Sameer Singh.
\newblock Autoprompt: Eliciting knowledge from language models with automatically generated prompts.
\newblock \emph{arXiv preprint arXiv:2010.15980}, 2020.

\bibitem[Sun et~al.(2022)Sun, Hu, and Saenko]{sun2022dualcoop}
Ximeng Sun, Ping Hu, and Kate Saenko.
\newblock Dualcoop: Fast adaptation to multi-label recognition with limited annotations.
\newblock \emph{Advances in Neural Information Processing Systems}, 35:\penalty0 30569--30582, 2022.

\bibitem[Sun et~al.(2021)Sun, Guo, and Li]{sun2021react}
Yiyou Sun, Chuan Guo, and Yixuan Li.
\newblock React: Out-of-distribution detection with rectified activations.
\newblock \emph{Advances in Neural Information Processing Systems}, 34:\penalty0 144--157, 2021.

\bibitem[Tian et~al.(2022)Tian, Liu, Pang, Liu, Chen, and Carneiro]{tian2022pixel}
Yu Tian, Yuyuan Liu, Guansong Pang, Fengbei Liu, Yuanhong Chen, and Gustavo Carneiro.
\newblock Pixel-wise energy-biased abstention learning for anomaly segmentation on complex urban driving scenes.
\newblock In \emph{European Conference on Computer Vision}, pages 246--263. Springer, 2022.

\bibitem[Van~Horn et~al.(2018)Van~Horn, Mac~Aodha, Song, Cui, Sun, Shepard, Adam, Perona, and Belongie]{van2018inaturalist}
Grant Van~Horn, Oisin Mac~Aodha, Yang Song, Yin Cui, Chen Sun, Alex Shepard, Hartwig Adam, Pietro Perona, and Serge Belongie.
\newblock The inaturalist species classification and detection dataset.
\newblock In \emph{Proceedings of the IEEE conference on computer vision and pattern recognition}, pages 8769--8778, 2018.

\bibitem[Vaswani et~al.(2017)Vaswani, Shazeer, Parmar, Uszkoreit, Jones, Gomez, Kaiser, and Polosukhin]{vaswani2017attention}
Ashish Vaswani, Noam Shazeer, Niki Parmar, Jakob Uszkoreit, Llion Jones, Aidan~N Gomez, {\L}ukasz Kaiser, and Illia Polosukhin.
\newblock Attention is all you need.
\newblock \emph{Advances in neural information processing systems}, 30, 2017.

\bibitem[Wang et~al.(2022)Wang, Li, Feng, and Zhang]{9879414}
Haoqi Wang, Zhizhong Li, Litong Feng, and Wayne Zhang.
\newblock Vim: Out-of-distribution with virtual-logit matching.
\newblock In \emph{2022 IEEE/CVF Conference on Computer Vision and Pattern Recognition (CVPR)}, pages 4911--4920, 2022.

\bibitem[Wang et~al.(2023)Wang, Li, Yao, and Li]{wang2023clipn}
Hualiang Wang, Yi Li, Huifeng Yao, and Xiaomeng Li.
\newblock Clipn for zero-shot ood detection: Teaching clip to say no.
\newblock In \emph{Proceedings of the IEEE/CVF International Conference on Computer Vision}, pages 1802--1812, 2023.

\bibitem[Wu et~al.(2023)Wu, Zhou, Pang, Zhou, Yan, Wang, and Zhang]{wu2023vadclip}
Peng Wu, Xuerong Zhou, Guansong Pang, Lingru Zhou, Qingsen Yan, Peng Wang, and Yanning Zhang.
\newblock Vadclip: Adapting vision-language models for weakly supervised video anomaly detection.
\newblock \emph{arXiv preprint arXiv:2308.11681}, 2023.

\bibitem[Xiao et~al.(2010)Xiao, Hays, Ehinger, Oliva, and Torralba]{xiao2010sun}
Jianxiong Xiao, James Hays, Krista~A Ehinger, Aude Oliva, and Antonio Torralba.
\newblock Sun database: Large-scale scene recognition from abbey to zoo.
\newblock In \emph{2010 IEEE computer society conference on computer vision and pattern recognition}, pages 3485--3492. IEEE, 2010.

\bibitem[Zhang et~al.(2023)Zhang, Rao, and Agrawala]{zhang2023adding}
Lvmin Zhang, Anyi Rao, and Maneesh Agrawala.
\newblock Adding conditional control to text-to-image diffusion models.
\newblock In \emph{Proceedings of the IEEE/CVF International Conference on Computer Vision}, pages 3836--3847, 2023.

\bibitem[Zhong et~al.(2021)Zhong, Friedman, and Chen]{zhong2021factual}
Zexuan Zhong, Dan Friedman, and Danqi Chen.
\newblock Factual probing is [mask]: Learning vs. learning to recall.
\newblock \emph{arXiv preprint arXiv:2104.05240}, 2021.

\bibitem[Zhou et~al.(2017)Zhou, Lapedriza, Khosla, Oliva, and Torralba]{zhou2017places}
Bolei Zhou, Agata Lapedriza, Aditya Khosla, Aude Oliva, and Antonio Torralba.
\newblock Places: A 10 million image database for scene recognition.
\newblock \emph{IEEE transactions on pattern analysis and machine intelligence}, 40\penalty0 (6):\penalty0 1452--1464, 2017.

\bibitem[Zhou et~al.(2021)Zhou, Ye, and Zhan]{zhou2021learning}
Da-Wei Zhou, Han-Jia Ye, and De-Chuan Zhan.
\newblock Learning placeholders for open-set recognition.
\newblock In \emph{Proceedings of the IEEE/CVF conference on computer vision and pattern recognition}, pages 4401--4410, 2021.

\bibitem[Zhou et~al.(2022{\natexlab{a}})Zhou, Yang, Loy, and Liu]{zhou2022conditional}
Kaiyang Zhou, Jingkang Yang, Chen~Change Loy, and Ziwei Liu.
\newblock Conditional prompt learning for vision-language models.
\newblock In \emph{Proceedings of the IEEE/CVF Conference on Computer Vision and Pattern Recognition}, pages 16816--16825, 2022{\natexlab{a}}.

\bibitem[Zhou et~al.(2022{\natexlab{b}})Zhou, Yang, Loy, and Liu]{zhou2022learning}
Kaiyang Zhou, Jingkang Yang, Chen~Change Loy, and Ziwei Liu.
\newblock Learning to prompt for vision-language models.
\newblock \emph{International Journal of Computer Vision}, 130\penalty0 (9):\penalty0 2337--2348, 2022{\natexlab{b}}.

\bibitem[Zhou et~al.(2023)Zhou, Pang, Tian, He, and Chen]{zhou2023anomalyclip}
Qihang Zhou, Guansong Pang, Yu Tian, Shibo He, and Jiming Chen.
\newblock Anomalyclip: Object-agnostic prompt learning for zero-shot anomaly detection.
\newblock \emph{arXiv preprint arXiv:2310.18961}, 2023.

\bibitem[Zhu and Pang(2024)]{zhu2024toward}
Jiawen Zhu and Guansong Pang.
\newblock Toward generalist anomaly detection via in-context residual learning with few-shot sample prompts.
\newblock \emph{arXiv preprint arXiv:2403.06495}, 2024.

\end{thebibliography}
}

\clearpage
\appendix
\twocolumn[
\begin{@twocolumnfalse}
\section*{\centering{\Large{Appendix}\\[40pt]}}
\end{@twocolumnfalse}
]

\section{Ablation Study of Hyperparameters $\beta$, $\gamma$}
As discussed in Section 3.2.1, the overall loss of our method comprises three components:
\begin{equation}
    \mathcal{L}_{Negative Prompts} = \mathcal{L}_{NIS} + \beta * \mathcal{L}_{NPD} + \gamma * \mathcal{L}_{NND},
\end{equation}
where $\beta$ and $\gamma$ are two hyperparameters.

 We conducted ablation studies on the values of \(\beta\) and \(\gamma\), with the results presented in the table \ref{tab:ablation_hyper}. It is observed that 1) setting either \(\gamma\) or \(\beta\) to zero, \ie, removing $\mathcal{L}_{NPD}$ or $\mathcal{L}_{NND}$, can lead to a significant decrease in AUROC and/or FPR95; and 2) our method achieves the best performance when \(\beta\) and \(\gamma\) are set to 0.1 and 0.05, respectively, indicating a greater importance of $\mathcal{L}_{NPD}$ than $\mathcal{L}_{NND}$.
\begin{table*}[!hb]
    \centering
    \scalebox{0.9}{
    \begin{tabular}{cc cc  cc   cc  cc |cc}
        \hline
        \hline 
        \specialrule{0em}{1pt}{1pt}
        \multirow{2}*{\textbf{$\beta$}} & \multirow{2}*{\textbf{$\gamma$}} & \multicolumn{2}{c}{\textbf{Texture}} &\multicolumn{2}{c}{\textbf{iNaturalist}} &\multicolumn{2}{c}{\textbf{Places}} & \multicolumn{2}{c|}{\textbf{SUN}} & \multicolumn{2}{c}{\textbf{Avg}}\\
        ~ & ~ &  AUC $\uparrow$  & FPR95 $\downarrow$ &  AUC $\uparrow$  & FPR95 $\downarrow$ &  AUC $\uparrow$  & FPR95 $\downarrow$ &  AUC $\uparrow$  & FPR95 $\downarrow$ &  AUC $\uparrow$  & FPR95 $\downarrow$\\
        \hline
        \textit{ablation of $\beta$}& ~ & ~& ~& ~& ~& ~& ~& ~& ~& ~\\
         0    &  0.05 & 89.35 & 37.73 & 96.85 & 11.65 & 90.50 & 32.86 & 89.48 & 26.01 & 91.55 & 27.06 \\
        0.01 &  0.05 & 89.86 & 36.64 & 98.53 & 8.23  & 91.88 & 29.13 & 93.3  & 24.04 & 93.39 & 24.51 \\
         0.05 &  0.05 & \textbf{91.74} & 36.12 & 97.15 & 7.23  & 92.67 & 28.17 & 94.36 & 23.24 & 93.98 & 23.69 \\
         0.2  &  0.05 & 90.94 & 35.41 & 98.27 & 7.16  & 92.61 & 27.80 & 93.66 & 22.93 & 93.87 & 23.33 \\
         0.5  &  0.05 & 90.69 & 36.21 & 97.79 & 8.73  & 91.58 & 28.32 & 92.85 & 25.84 & 93.23 & 24.78 \\
         1    &  0.05 & 90.39 & 36.45 & 97.13 & 10.27 & 91.63 & 30.92 & 92.55 & 25.55 & 92.93 & 25.80 \\

        \hline
        ~ & \textit{ablation of $\gamma$}& ~ & ~& ~& ~& ~& ~& ~& ~& ~& ~\\
         0.1  &  0    & 89.52 & 37.11 & 98.02 & 10.86 & 91.4  & 33.08 & 92.6  & 25.08 & 92.89 & 26.53 \\
         0.1  &  0.01 & 89.87 & 36.13 & 98.08 & 9.04  & 92.44 & 29.14 & 93.44 & 24.85 & 93.46 & 24.79 \\
         0.1  &  0.1  & 91.56 & 35.84 & 98.25 & 7.36  & 93.17 & 27.63 & 95.28 & 23.33 & 94.57 & 23.54 \\
         0.1  &  0.2  & 90.87 & 35.41 & 97.36 & 7.89  & 92.15 & 28.40 & 93.42 & 22.97 & 93.45 & 23.67 \\
         0.1  &  0.5  & 91.35 & 36.39 & 98.18 & 9.05  & 92.35 & 30.29 & 92.14 & 25.91 & 93.51 & 25.41 \\
         0.1  &  1    & 90.77 & 37.37 & 98.40 & 9.46  & 91.23 & 31.77 & 92.84 & 23.98 & 93.31 & 25.65 \\
        \hline
        0.1 & 0.05 & 91.60	& \textbf{35.21}	 & \textbf{98.73}& \textbf{6.32}	& \textbf{93.34}	& \textbf{27.60}	& \textbf{95.55}	& \textbf{22.89}	& \textbf{94.81}	& \textbf{23.01}\\
    \end{tabular}
    }
    \caption{Ablation experiments for hyperparameters \(\beta\) and \(\gamma\). We fixed the values of \(\beta\) and \(\gamma\) at 0.1 and 0.05, respectively, and conducted controlled experiments. The best results are highlighted in \textbf{bold}, and each result was averaged over three trials.}
    \label{tab:ablation_hyper}
\end{table*}
\section{Dataset Information}
In this section, we provide a detailed description of the datasets used. 

For the conventional OOD detection, the ID images and OOD images are from different datasets: for ID, we utilized ImageNet-1k as our in-distribution datasets, consisting of 1,000 categories with 1,281,167 training images and 50,000 validation images. We train with 16 images per class for few-shot learning and employ all validation images as ID images for testing. Regarding OOD datasets, we followed \cite{ming2022delving, miyai2023locoop, wang2023clipn}, employing Texture~\cite{cimpoi2014describing}, iNaturalist~\cite{van2018inaturalist}, Places~\cite{zhou2017places} and SUN~\cite{xiao2010sun} as our OOD test datasets. Further details of these OOD test datasets are provided below.

\noindent\textbf{Texture.}
Describable Textures Dataset~\cite{cimpoi2014describing} contains images of textures and abstract patterns. As no categories overlap with ImageNet-1k, we use the entire dataset as OOD images.

\noindent\textbf{iNaturalist.}
Containing images from the real world, iNaturalist~\cite{van2018inaturalist} has 13 super-categories and 5,089 subcategories covering plants, insects, birds, mammals, and so on. We use the subset that contains 110 plant classes not overlapping with ImageNet-1k.

\noindent\textbf{Places365.}
As a large scene photograph dataset, Places365~\cite{zhou2017places} contains photos that are labeled with scene semantic categories from three macro-classes: Indoor, Nature, and Urban. The subset we use is sampled from 50 categories that are not present in ImageNet-1k.

\noindent\textbf{SUN.}
Scene Understanding Dataset~\cite{xiao2010sun} contains 899 categories that cover more than indoor, urban, and natural places with or without human beings appearing. We use the subset which contains 50 natural objects not showing in ImageNet-1k.

Regarding hard Out-of-Distribution detection, both the ID and OOD classes are derived from ImageNet-1k. We adhered to the three splits proposed in ~\cite{palechor2023large} and introduce another split: designating the first 100 classes of ImageNet as ID and the remaining 900 classes as OOD, which aims to facilitate comprehensive OOD detection across the entire ImageNet-1k dataset. The number of ID and OOD categories in each split is illustrated in Table \ref{tab:classes_num_information}. 
\begin{table}[!ht]
\centering
\scalebox{0.8}{
\begin{tabular}{@{}lll@{}}
\toprule
& \textbf{ID}          & \textbf{OOD}               \\ 
        \midrule
\multirow{2}*{\textbf{Split-1}}              & All dog classes               & Non-animal classes             \\
~ & 116:\ 1856 / 5800 & 166:\ ---\ /\ 8300\\
                        
\multirow{2}*{\textbf{Split-2}}            & Half of hunting dog classes       & Other 4-legged animal classes   \\
~ & 30:\ 480 / 1500 & 55:\ ---\ /\ 2750\\
\multirow{2}*{\textbf{Split-3}}             & Mix of common classes  & Mix of common classes \\
~ & 151:\ 2416 / 7550 & 164:\ ---\ /\ 8200\\
\multirow{2}*{\textbf{Split-4}}              & First 100 classes   & Remaining 900 classes \\
~ & 100:\  1600 / 5000 & 900:\ ---\ /\ 45000\\
 \bottomrule
\end{tabular}
}
\caption{All ImageNet-1K splits for hard OOD detection. Given are the numbers of \textit{classes}\ :\ \textit{training}\ /\ \textit{test} samples.}
\label{tab:classes_num_information}
\end{table}

\section{Implementation Details}
All methods are implemented in Pytorch 1.10. We run all OOD detection experiments on an NVIDIA V100 GPU. For CLIP, we follow LoCoOp~\cite{miyai2023locoop} and employ the implementation of OpenClip~\cite{ilharco_gabriel_2021_5143773}, using parameters pre-trained with LAION-2B~\cite{schuhmann2022laionb}. A two-stage training method is adopted: during the first stage, we train only positive prompts, following the same training scheme as CoOp~\cite{zhou2022learning}. A single positive prompt is trained for all classes, with the number of context tokens set to 16. In the second stage, we freeze the positive prompts and initialize the negative prompt parameters as the positive prompt, followed by training for 15 epochs. During the testing phase, we employ the MCM~\cite{ming2022delving}, where the highest similarity between the positive text feature and the image feature, post-Softmax of all similarities, is taken as the result.

\section{Computational Time}
A comparative analysis of the computational time for prompt learning-based approaches on ImageNet-1k/Texture is also performed. As indicated in Table \ref{tab:computation_time}, our approach has a higher training overhead compared to CoOp, yet it is much lower than that of LoCoOp. Regarding the inference time, NegPrompt aligns with CoOp and surpasses LoCoOp. 
\begin{table}[!ht]
\centering
\scalebox{1}{
\begin{tabular}{@{}lll@{}}
\toprule
Method &   \textbf{Training Time}          & \textbf{Inference Time}               \\ 
        \midrule
\textbf{LoCoOp}~\cite{miyai2023locoop} &  475 min & 20 min \\
\textbf{CoOp}~\cite{zhou2022learning} & 341 min & 3 min \\
\textbf{NegPrompt}(ours) & 443 min & 3 min\\
 \bottomrule
\end{tabular}
}
\caption{The computation time of prompt learning-based methods. For LoCoOp~\cite{miyai2023locoop}, we use its official implementation. The training time and inference time are averaged from three separate experiments.}
\label{tab:computation_time}

\end{table}

\end{document}